\documentclass[10pt,twocolumn,letterpaper]{article}

\usepackage{cvpr}              
\usepackage{newfloat}
\usepackage{listings}
\usepackage{algorithm}
\usepackage{algorithmic}

\usepackage{amsmath}
\usepackage{multirow}
\usepackage{pifont}

\usepackage[accsupp]{axessibility}

\definecolor{cvprblue}{rgb}{0.21,0.49,0.74}
\usepackage[pagebackref,breaklinks,colorlinks,allcolors=cvprblue]{hyperref}

\newcommand{\cmark}{\textcolor{ForestGreen}{\ding{51}}}%
\newcommand{\xmark}{\textcolor{red}{\ding{55}}}%
\newcommand{\omark}{\textcolor{RoyalBlue}{\ding{108}}} 

\title{What’s Wrong with Synthetic Data for Scene Text Recognition? \\  

A Strong Synthetic Engine with Diverse Simulations and Self-Evolution

}

\author{
  Xingsong Ye$^{1,2}$\thanks{Equal contribution.},
  Yongkun Du$^{1,2}$\footnotemark[1],
  JiaXin Zhang$^{3}$,
  Chen Li$^{3}$,
  Jing LYU$^{3}$,
  Zhineng Chen$^{1,2}$\thanks{Corresponding author. E-mail: zhinchen@fudan.edu.cn}
\\
$^1$Institute of Trustworthy Embodied AI, Fudan University \\
$^2$Shanghai Key Laboratory of Multimodal Embodied AI \hfill
$^3$WeChat Vision, Tencent Inc.\\
}

\begin{document}
\maketitle

\begin{abstract}
Large-scale and categorical-balanced text data is essential for training effective Scene Text Recognition (STR) models, which is hard to achieve when collecting real data. Synthetic data offers a cost-effective and perfectly labeled alternative. However, its performance often lags behind, revealing a significant domain gap between real and current synthetic data. In this work, we systematically analyze mainstream rendering-based synthetic datasets and identify their key limitations: insufficient diversity in corpus, font, and layout, which restricts their realism in complex scenarios. To address these issues, we introduce \textbf{UnionST}, a strong data engine synthesizes text covering a union of challenging samples and better aligns with the complexity observed in the wild. We then construct \textbf{UnionST-S}, a large-scale synthetic dataset with improved simulations in challenging scenarios. Furthermore, we develop a self-evolution learning (SEL) framework for effective real data annotation. Experiments show that models trained on UnionST-S achieve significant improvements over existing synthetic datasets. They even surpass real-data performance in certain scenarios. Moreover, when using SEL, the trained models achieve competitive performance by only seeing 9\% of real data labels.
Code is available at \url{https://github.com/YesianRohn/UnionST}.

\end{abstract}

\section{Introduction}

\begin{figure}[t]
\centering
\includegraphics[width=0.45\textwidth]{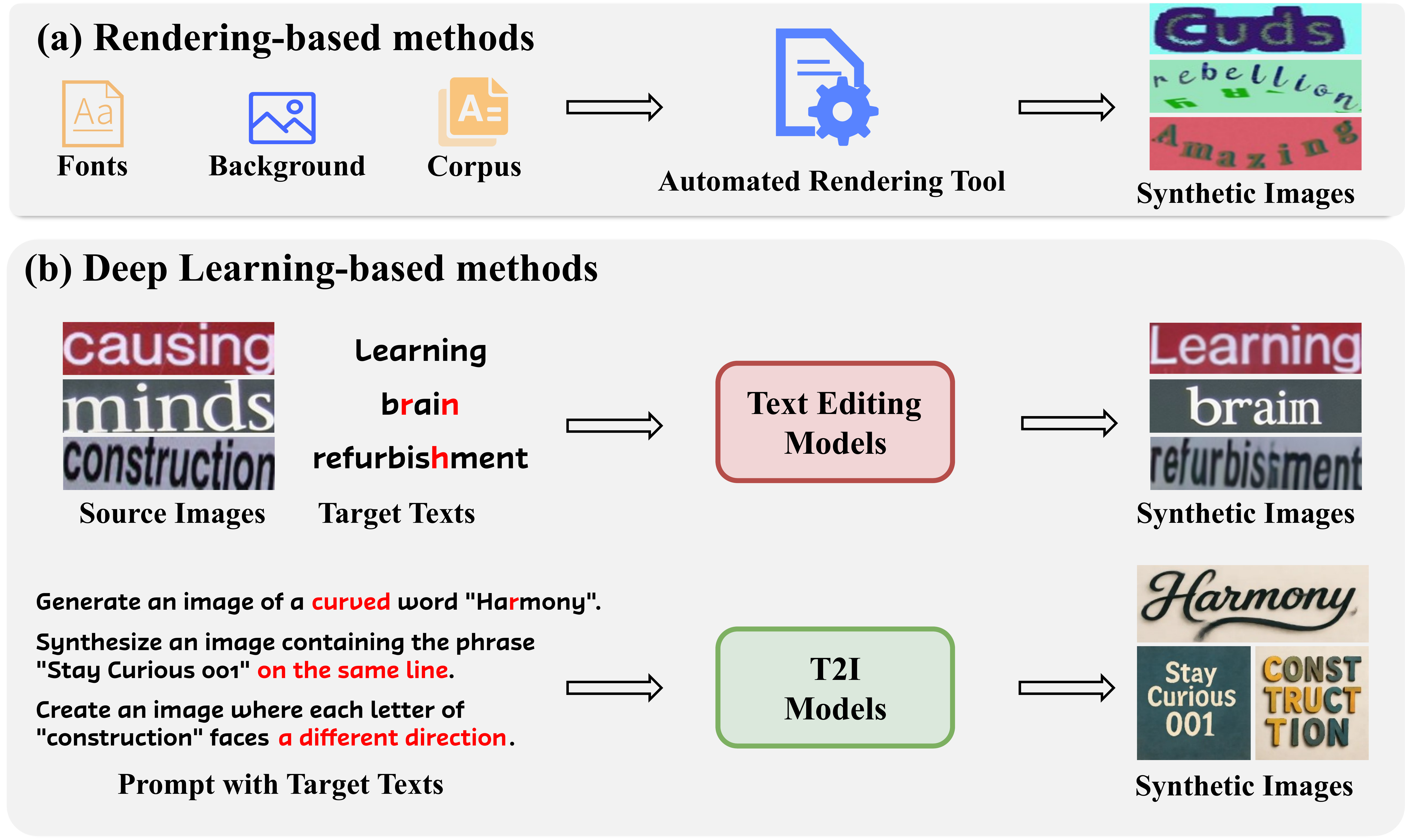} 
\caption{Comparison of two scene text data synthesis paradigms. ``T2I'' stands for Text-to-Image and \textcolor{red}{red} highlights indicate that the edited or rendered images do not meet the specified condition.}  
\label{fig:synth_comp}
\vspace{-10pt}
\end{figure}

Scene Text Recognition (STR) aims to precisely extract text from natural images that feature intricate backgrounds and a variety of imaging conditions. Beyond traditional Optical Character Recognition (OCR) applications, this task has become increasingly vital in the era of large language models (LLMs), as it contributes substantial training data for corpus construction during pre-training. STR is highly challenging due to factors such as varying shooting angles, background clutter, severe noise and blur, and the wide variability in text styles.  A key driver of STR is the quantity and quality of training data, which has experienced three main stages: (1) early small-scale real-annotated datasets~\cite{Wang2011SVT, Strokelets}, (2) large-scale synthetic datasets~\cite{shi2017crnn, qiao2020seed, zheng2024cdistnet}, and (3) the current stage that combines large-scale real and synthetic data~\cite{rang2024empirical, du2024svtrv2, yang2025ipad}. There are two paradigms for synthetic data generation, i.e., traditional rendering-based methods~\cite{mj, st, yim2021synthtiger, liao2020synthtext3d, long2020unrealtext} and deep learning-based generation approaches~\cite{wu2019editing, zeng2024textctrl, zhu2023conditional, textssr}.  While the former ensures ground-truth labels, the latter aims to produce data that is visually more realistic and closer to the distribution of texts in the wild. However, recent researches~\cite{scenevtg, textssr} indicate that most generative models remain at the stage of aesthetically pleasing text creation, rather than accurate and realistic text synthesis at scale. As shown in Fig.~\ref{fig:synth_comp}, when generating scene text training data, rendering-based methods consistently surpass popular generative models in accuracy and controllability. Moreover, the rendering-based methods are cheaper because they rely on CPUs, costing only 1/20 of diffusion-based TextSSR~\cite{textssr} and 1/10,000 of the closed-source Nano Banana.

Although several large-scale real datasets \cite{jiang2023revisiting} have been publicly released to date, the synthesis of training data is still of vital importance for STR. 
Recent studies reveal that even state-of-the-art (SOTA) models \cite{cppd, du2024svtrv2, du2025igtr} trained on the largest available real datasets still leave substantial room for improvement. This indicates that STR remains unsolved from the data perspective and highlights the necessity of integrating high-quality synthetic text to overcome the current accuracy bottleneck. However, the performance of existing synthetic data often lags far behind that of much smaller real datasets~\cite{baek2021if, whatwrong, rang2024empirical}. This exposes a fundamental issue: a substantial domain gap exists between synthetic and real-world data~\cite{cppd}. These findings motivate us to revisit two critical questions: \textit{(1) Is the potential of synthetic data exhausted, and how can rendering-based engines be further improved? (2) Can enhanced synthetic data boost the performance of existing labeled real data, or be used with unlabeled real data for self-evolution to reduce annotation costs, perhaps even achieving performance comparable to labeled data?}

\begin{figure}[t]
\centering
\includegraphics[width=0.45\textwidth]{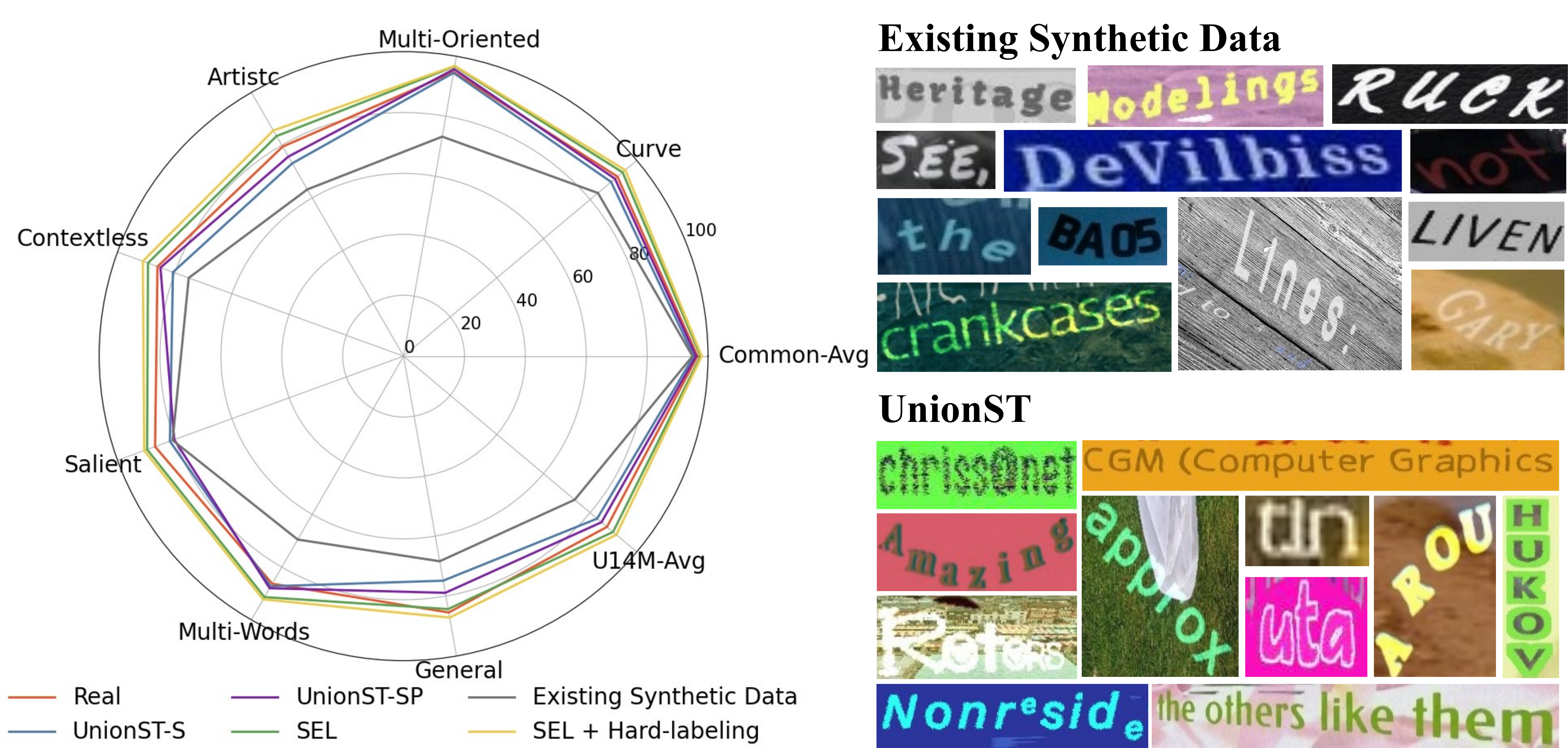} 
\caption{\textbf{Left:} Quantitative comparisons across multiple scenarios, including common and seven challenging cases, are performed based on the normalized accuracy rate (\%). \textbf{Right:} Illustrative examples of traditional synthetic engines (MJ~\cite{mj}, ST~\cite{st}, CurvedST~\cite{curvest}, SynthAdd~\cite{li2019show}, and SynthTIGER~\cite{yim2021synthtiger}) and our UnionST, which provides more diverse (various text layouts and content), realistic, and challenging samples. }  
\label{fig:radar}
\vspace{-10pt}
\end{figure}

For the first question, we systematically evaluate mainstream rendering-based synthetic datasets, totaling 36M samples. The quantitative results and qualitative visualizations in Fig.~\ref{fig:radar} demonstrate that samples from them exhibit the following gaps compared to real data:
(1) Most synthetic samples consist of single, semantically rich, and short words, which leads to poor performance in multi-words and contextless scenarios. (2) Their fonts are conventional and easily recognizable, resulting in suboptimal performance on artistic text. (3) The layouts are typically simple and monotonous, with all characters arranged horizontally and uniformly sized. This simplicity prevents the synthesis of curved or multi-oriented text, further limiting the diversity of the generated data. Consequently, STR models trained on such datasets perform poorly in challenging scenarios.

To better simulate the appearance of text in the wild, we propose \textbf{UnionST}, a strong synthetic data engine that emphasizes generating text images covering a union of challenging scenarios. UnionST models complex factors by enriching both the corpus and font diversity, as well as redefining target text placement, thereby narrowing the gap between synthetic and real-world data.
Leveraging UnionST, we construct \textbf{UnionST-S}, a large-scale synthetic dataset tailored for diverse and challenging scenarios. Experiments demonstrate that training solely on UnionST-S significantly outperforms traditional synthetic datasets. Notably, on specific subsets such as Multi-Words, UnionST-S even surpasses real dataset performance. These results demonstrate that synthetic data can still play a greater role, particularly when real annotations are unavailable.

We illustrate UnionST's practical impact through the following scenarios: (1) Near Real Results with Minimal Annotation. Fine-tuning a UnionST-S pre-trained model using only 1\% (32K) randomly sampled real annotations yields performance on par with training on the entire real dataset (3.2M). (2) Unlocking the Potential of Unlabeled Real Data. We introduce a self-evolution learning (SEL) framework built on UnionST-S: (a) Pseudo-label large-scale unlabeled real data and use the predicted texts as ``corpus'' to construct a new dataset, UnionST-P (5M). (b) Then train the STR model on the combined datasets (UnionST-SP, 10M) and, iteratively pseudo-label high-confidence (e.g., $\geq 0.9$) unlabeled samples and fine-tune the model. After two rounds, the model achieves 89.81\% average accuracy on Union14M-Benchmark, outperforming models trained on the full real dataset by 2.59\%. (3) Achieving SOTA Performance. We further fine-tune the UnionST-SP pre-trained model on real data, obtaining a new SOTA average accuracy of 91.39\% on Union14M-Benchmark. (4) Reducing Annotation Cost. Continuing the SEL, only the remaining low-confidence but challenging samples ($ \sim $ 9\%) require manual annotation. Only fine-tuning on them narrows the performance gap to within 0.16\% of the above SOTA result, while reducing manual annotation requirements by 91\%.

In summary, our main contributions are as follows:
\begin{itemize}
    \item We compare two synthetic data paradigms and note that rendering-based methods remain useful but perform poorly in challenging scenarios, as they fail to fully simulate and combine complex conditions.  Therefore, the potential of synthetic data has not been fully tapped.
    \item We propose \textbf{UnionST}, a strong synthetic data engine with diverse simulations. Based on this, we release \textbf{UnionST-S}, which significantly outperforms traditional synthetic datasets, and \textbf{UnionST-P}, leveraging pseudo-corpus to approach real-data performance.
    \item The SEL enabled by UnionST substantially reduces the annotation workload while maintaining an accuracy comparable to that achieved with full supervision after UnionST pre-training.
\end{itemize}

\section{Related Work}
\subsection{Data Synthesis for STR}
Early synthetic data for STR followed the rendering-based paradigm first defined by MJ~\cite{mj}. This approach includes font rendering, border and shadow processing, background coloring, composition, projective distortions, blending with real images, and adding noise. ST~\cite{st} improves MJ by considering depth and segmentation, enabling text placement on more suitable surfaces. Later methods like SynthAdd~\cite{li2019show} and CurvedST~\cite{curvest} add support for special characters and simulate curved text scenarios. SynthTIGER~\cite{yim2021synthtiger} offers a unified framework, analyzing the impact of each synthetic component and combining effects from MJ and ST. 3D-based methods have also emerged. SynthText3D~\cite{liao2020synthtext3d} and UnrealText~\cite{long2020unrealtext} insert 2D text into 3D environments for realism. More recent work leverages deep learning for data synthesis. Scene text editing methods \cite{wu2019editing, mostel, zeng2024textctrl} aim to expand data by editing text within specified regions. Additionally, some works utilize diffusion models for data synthesis directly~\cite{zhu2023conditional, scenevtg, textssr}. Although these generative methods create more realistic examples, they often struggle with accuracy and fine control in complex scenarios.

\subsection{Data Perspective Analysis of STR}
TRBA~\cite{whatwrong} compares STR models trained on MJ+ST and highlight several challenges, such as handling calligraphic fonts, vertical texts, and low-resolution images. To address the lack of real annotations, STR-Fewer-Labels~\cite{baek2021if} shows that using only 1.7\% of real data, together with RandAugment~\cite{cubuk2020randaugment} and semi-supervised learning, can match full MJ+ST performance. Union14M~\cite{jiang2023revisiting} proposes challenging benchmarks and stresses that current STR issues are far from being resolved. CCD~\cite{ccd}, CCDPlus~\cite{ccdplus}, and ViSu~\cite{qu2024boosting} further leverage synthetic data to drive self-/semi-supervised learning, thereby reducing annotation costs while improving model performance.

\section{Methodology}

\subsection{Synthetic Data Engine}

Rendering-based scene text synthesis is a popular approach for data generation, as it guarantees perfectly accurate labels and enables large-scale synthesis with ease. However, existing methods often fail to generate samples that capture the challenging and diverse conditions. In essence, the pipeline of rendering-based engines involves selecting the target text and font, rendering the visual text, and placing it onto a suitable background region. To understand where the current synthesis engine fails, we need to carefully examine and improve each component (see Tab.~\ref{tab:comparison_components}).

\noindent\textbf{Enrich the Corpus.} Most existing corpora are composed of isolated semantic words, and such data is insufficient to capture the complexity, diversity, and reality of real-world situations. We augment the corpus with additional text types designed to better reflect practical challenges:
\begin{itemize}
    \item \textbf{Contextless}:
    To simulate text instances lacking semantic context, like license plates or phone numbers, we generate random characters to form texts.
    \item \textbf{Incomplete}:
    Real-world images often contain partially occluded or cropped text, resulting in missing characters at the beginning, end, or within words. To mimic this, we create augmented samples by randomly removing one character from different positions within semantic words.
    \item \textbf{Multi-Words}:
    Scene text is frequently composed of multiple words or entire phrases, rather than a single word. To capture this phenomenon, we incorporate common phrases, concatenated words, and text fragments of varying lengths extracted from natural language corpora.
\end{itemize}

\begin{table}[t]
\centering
\resizebox{\linewidth}{!}{%
\begin{tabular}{c|cccc|c|cccc}
\toprule
\multirow{2}{*}{\textbf{Method}} & \multicolumn{4}{c|}{\textbf{Corpus}} & \multicolumn{1}{c|}{\textbf{Font}} & \multicolumn{4}{c}{\textbf{Layout}} \\
  & Common & Contextless & Incomplete & Multi-Words &  Amount & Curve & Multi-Oriented & Multi-Sized & Salient  \\
\midrule
MJ~\cite{mj} & \cmark & \xmark & \xmark & \xmark & 1.4K & \xmark & \xmark & \xmark & \xmark \\
ST~\cite{st} & \cmark & \xmark & \xmark & \xmark & 1.2K & \xmark & \omark & \xmark & \omark \\
SynthAdd~\cite{li2019show} & \xmark & \cmark & \xmark & \xmark & 1.2K & \xmark & \xmark & \xmark & \omark \\
CurvedST~\cite{curvest} & \cmark & \xmark & \xmark & \xmark & 1.2K & \cmark & \omark & \xmark & \omark \\
SynthTIGER~\cite{yim2021synthtiger} & \cmark & \xmark & \xmark & \xmark & 3.6K & \cmark & \omark & \xmark & \cmark \\
UnrealText~\cite{long2020unrealtext} & \cmark & \xmark & \xmark & \xmark & 3.1K & \xmark & \omark & \xmark & \omark \\
\midrule
UnionST (Ours) & \cmark & \cmark & \cmark & \cmark & 113.8K & \cmark & \cmark & \cmark & \cmark \\
\bottomrule
\end{tabular}%
}
\caption{Comparison of component implementations in different STR data engines. ~\xmark~ indicates no implementation, ~\omark~ indicates partial or uncontrollable implementation, and ~\cmark~ indicates full and controllable implementation.}
\label{tab:comparison_components}
\end{table}

\noindent\textbf{Cover Various Fonts.} The diversity and accuracy of fonts play a crucial role in ensuring text variability and enhancing model performance.  To better mimic the realistic variety of typefaces, we collect a broad set of publicly available fonts and automatically filter out those with indistinguishable uppercase and lowercase glyphs, ensuring clear character distinction for robust training.

\begin{figure*}[t]
\centering
\includegraphics[width=0.8\textwidth]{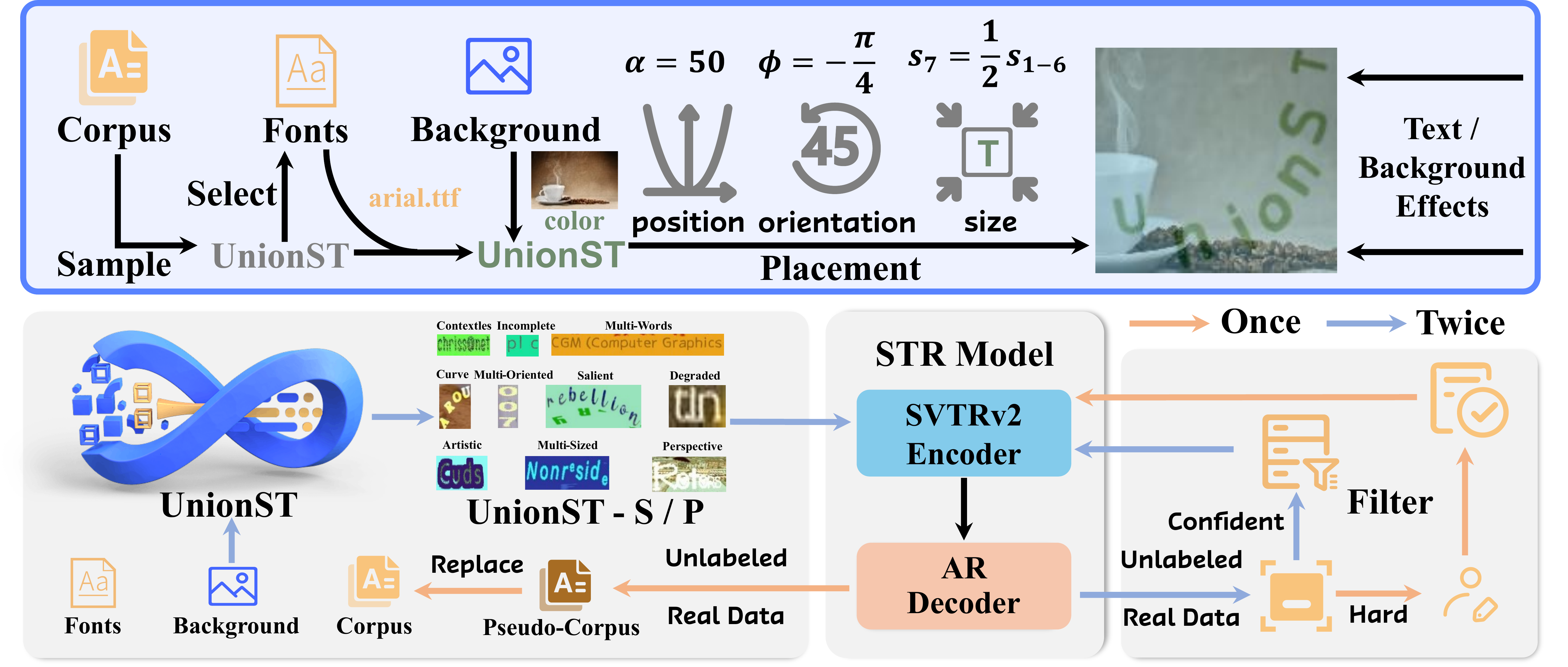} 
\caption{
Pipelines of the UnionST data engine (top) and our SEL  framework (bottom). \textbf{Top:} We randomly sample a text (e.g., \texttt{UnionST}) from the corpus, select a font (e.g., \texttt{arial.ttf}) that supports all characters, and render the text using this font in a color chosen from a predefined colormap. The rendered text is then placed onto a background with effects using our placement algorithm. For more UnionST’s data visualization, see Tab.~\ref{fig:data_view} in the supplementary. \textbf{Bottom:} The lines represent the number of passes. \underline{Left:} UnionST-S and UnionST-P come from different corpus, and UnionST-P is combined with UnionST-S for STR model retraining. \underline{Right:} Pseudo-labels undergo two rounds of self-iteration, followed by one round with manually annotated data. Each iteration fine-tunes the previous model. }
\label{fig:pipeline}
\end{figure*}

\noindent\textbf{Support Challenging Layouts.}
As illustrated in Fig.~\ref{fig:pipeline}, ``Curve'' and ``Multi-Oriented'' texts are often accompanied by inconsistencies in character size (``Multi-Sized''). To realistically capture such complex text layouts, we model the position, orientation, and size of each character independently and render them individually as separate layers. Specifically, for each character $i$ ($i=1,\ldots,N$) in the given text $T$, we define the placement as:
\begin{equation}
\text{placement}_T = \left\{ (p_i, o_i, s_i) \mid i = 1, \ldots, N \right\}
\end{equation}
where $p_i$ denotes the position, $o_i$ the orientation, and $s_i$ the size of the $i$-th character. The position $p_i$ is computed as:
\begin{equation}
p_i
=
\begin{pmatrix}
\cos\phi & -\sin\phi \\
\sin\phi & \cos\phi
\end{pmatrix}
\begin{pmatrix}
x_i \\
a x_i^2 + b 
\end{pmatrix}
\end{equation}
and the orientation $o_i$ is given by:
\begin{equation}
o_i = \arctan(2a x_i) + \phi
\end{equation}
Here, the curvature parameter $a$ is sampled from the range $[20, 200]$ to generate text layers with varying degrees of curvature, where $a=0$ corresponds to straight text and $b$ denotes the corresponding coordinate on the straight line. The global rotation angle $\phi$ is uniformly sampled from $[0, 2\pi)$ to introduce diverse orientations. For vertical text, it is equivalent to swapping the horizontal and vertical axes.

\noindent\textbf{UnionST Pipeline.}
Based on the above analysis and the proposed improvements, we design the UnionST engine with the following workflow (see the top of Fig.~\ref{fig:pipeline}): (a) We first randomly sample text from a prepared corpus and select a compatible font by querying the font’s character set.  (b) Each character is rendered individually as a separate layer, and we compute and placement parameters, including position, orientation, and size.  (c) We then apply various effects to the text, such as elastic deformation, perspective transformation, and border.  (d) Next, we randomly select a background and determine the text color using a predefined color correspondence table.  We further augment the background by adding random text instances, shadows, or embossing effects.  The final output is a synthesized image paired with its corresponding text label. We also explore online data augmentation during model training, which provides greater diversity and effectiveness. Specifically, to simulate small or blurry samples, we apply downsampling and transmission distortion augmentations (\textbf{DTAug}) during training.

Using this pipeline, we generate a large-scale synthetic dataset, \textbf{UnionST-S} (with 5M samples), which aims to provide highly realistic simulations of real-world scenarios, even in the absence of real data exposure.

\begin{table*}[t]\footnotesize
\centering
\setlength{\tabcolsep}{3pt}{
\begin{tabular}{c|c|c|ccccccc|cccccccc}
\toprule
\multirow{2}{*}{\textbf{Type}} &\multirow{2}{*}{\textbf{Model}} & \multirow{2}{*}{\textbf{Venue}} & \multicolumn{7}{c|}{\textbf{Common Benchmarks}}                                                                & \multicolumn{8}{c}{\textbf{Union14M-Benchmark}}                                             \\
 & & &
\textit{IIIT} & \textit{SVT}  & \textit{IC13} & \textit{IC15} & \textit{SVTP} & \textit{CUTE} & \textit{AVG} &
\textit{CUR} & \textit{MLO} & \textit{ART} & \textit{CTL} & \textit{SAL} & \textit{MLW} & \textit{GEN} & \textit{AVG}                  
\\
\midrule

\multirow{4}{*}{\textbf{CTC}} & CRNN~\cite{shi2017crnn}    &   TPAMI 2016    &92.90 &85.63 &91.83 &75.37 &76.12 &82.29 &84.02 &33.59 &64.21 &36.33 &49.29 &16.49 &65.78 &48.98 &44.95 \\
& ViT-CTC  & -  &   96.33 &	93.35 &	97.20 &	83.49 &	88.22 &	92.71 &	91.88 &	68.47	& 84.59 & 	58.00 &	71.12	& 52.24 &	79.61 &	65.71 &	68.53     \\
& SVTR~\cite{duijcai2022svtr}    &   IJCAI 2022    &97.50 &94.28 &97.78 &85.64 &90.08 &95.49 &93.46 &74.73 &86.56 &62.44 &74.45 &62.67 &81.80 &68.67 &73.05 \\
& SVTRv2~\cite{du2024svtrv2}    &   ICCV 2025     & \textbf{98.20} &	95.98 &	\textbf{98.60} &	87.85 &	94.26 &	\textbf{97.22} & 95.35  & 83.39	& 91.31	& 67.11	& 77.41	& 70.31 &	83.25	& 71.10 & 77.70 \\
\midrule
\multirow{4}{*}{\textbf{PD}} & ABINet~\cite{abinet}    &   CVPR 2021     &97.00 &93.35 &96.97 &85.04 &90.39 &94.10 &92.81 &73.04 &85.10 &58.44 &64.31 &59.25 &77.31 &63.98 &68.78 \\
& LPV~\cite{lpv}  & IJCAI 2023 & 97.77 &	95.98	& 96.85	& 86.47	& 90.85	& 94.44	& 93.73	& 78.44 &	89.85 &	65.00 &	71.63 &	66.65 &	81.31 &	69.25 &	74.59 \\
& BUSNet~\cite{busnet}  & AAAI 2024 & 97.20	& 95.52 &	96.62 &	85.75	& 91.94 &	94.10 &	93.52 &	73.62 &	87.66 &	63.44 &	69.58	& 62.22 &	79.61 &	68.70 &	72.12 \\
& CPPD~\cite{cppd} & TPAMI 2025 & 97.17	&95.36	&96.85	&85.15	&91.94	&95.49	&93.66	&83.80	&90.80	&67.11	&75.74	&70.06	&82.04	&69.82	&77.05\\
\midrule
\multirow{5}{*}{\textbf{AR}} & PARSeq~\cite{parseq}    &   ECCV 2022   &97.77 &94.74 &97.32 &85.64 &92.09 &93.75 &93.55 &73.54 &88.31 &64.89 &72.66 &68.48 &79.85 &70.24 &74.00    \\
& MAERec~\cite{jiang2023revisiting}    &   ICCV 2023     & 98.10 & \textbf{96.91} & 98.48 & 88.13 & \textbf{95.97} & 95.83 & \textbf{95.57} & 78.94 & 90.80 & 68.44 & 77.02 & 71.45 & 83.01 & 73.29 & 77.56 \\
& OTE~\cite{ote}    &   CVPR 2024  &97.77 &94.74 &97.32 &85.64 &92.09 &93.75 &93.55 &73.54 &88.31 &64.89 &72.66 &68.48 &79.85 &70.24 &74.00   \\
& SMTR~\cite{smtr}    &   AAAI 2025    &97.67 &95.52 &97.78 &86.36 &91.01 &94.10 &93.74 &80.87 &88.31 &65.33 &76.89 &69.24 &84.59 & 71.60 &76.69 \\
& SVTRv2-AR    &   -    & \textbf{98.20}  & 95.98 & 98.48 & \textbf{88.29} & 94.11 & 96.87 & 95.32 & \textbf{88.99} & \textbf{94.45} & \textbf{73.11} & \textbf{80.62} & \textbf{81.68} & \textbf{87.26} & \textbf{74.87} & \textbf{83.00} \\

\bottomrule
\end{tabular}}
\caption{Quantitative comparison of different  STR models trained on the 5M UnionST-S dataset. The values are all percentages (\%), and AVG represents the arithmetic mean. The supplementary further provides comparisons and analyses with the real data.}
\label{tab:comp_str}
\end{table*}

\subsection{STR Model Selection}

STR models rely heavily on large-scale, diverse datasets for effective training, making advances in data generation crucial for improving recognition performance. To maintain a focus on data-centric challenges, we adopt the SOTA STR model SVTRv2~\cite{du2024svtrv2} and modify it to an autoregressive (AR) architecture. Specifically, we replace the original CTC-based~\cite{ctc} decoder in SVTRv2 with an attention-based AR decoder~\cite{NIPS2017_attn, Sheng2019nrtr}. SVTRv2 surpasses previous encoder-decoder approaches even when trained with only the CTC decoder. While CTC-based methods work well in normal situations, they struggle with highly curved or multi-oriented text due to their assumption of monotonic character alignment. In contrast, attention-based decoders offer flexible alignment between input and output sequences, enabling more robust recognition in such challenging cases. This modification also allows us to fully leverage the potential of our synthetic dataset.

\subsection{Self-Evolution Learning}
\label{sec:3-3}


\noindent\textbf{Pseudo-labeling for Corpus Augmentation.}
Although UnionST-S comes from a comprehensive and universal corpus, there is still a gap between it and the actual text distribution. To further bridge this gap between real and synthetic data, we propose a generation strategy to better approximate the real data distribution:
\begin{enumerate}
    \item Utilize model $\mathcal{M}_a$ (e.g., SVTRv2-AR) trained on UnionST-S to generate pseudo-labels $\hat{\mathcal{Y}}_U$ for a large-scale unlabeled real dataset $\mathcal{D}_U$.
    \item Employ $\hat{\mathcal{Y}}_U$ as the target corpus to synthesize a new dataset, \textbf{UnionST-P} (comprising 5M samples), using the UnionST engine.
    \item Combine UnionST-S and UnionST-P to construct the augmented dataset \textbf{UnionST-SP}, resulting in 10M samples for retraining model $\mathcal{M}_b$.
\end{enumerate}

\noindent\textbf{Iterative Self-Refinement (ISR).}
The ISR scheme incrementally augments the training set with high-confidence pseudo-labeled samples, while reserving manual annotation exclusively for the most challenging instances. High-confidence pseudo-labeled samples serve as ``approximately correct'' labels, effectively expanding the training quantities and distribution. Low-confidence samples, often representing hard cases, are prioritized for manual annotation due to their higher marginal value. Through multiple rounds of self-refinement, model performance approaches the upper bound achievable with full supervision, while drastically reducing the amount of manual labeling required.

\vspace{-0.5em}
\begin{algorithm}[h]
\caption{Iterative Self-Refinement}
\begin{algorithmic}[1]
\STATE Initialize model $\mathcal{M}_0 = \mathcal{M}_b$ 
\FOR{$t=1$ to $T$}
    \STATE Use $\mathcal{M}_{t-1}$ to recognize on unlabeled data $\mathcal{D}_U$, select pseudo-labeled samples $\mathcal{D}_P^{(t)}$ with confidence above threshold $\tau$
    \STATE Fine-tune on $ \mathcal{D}_P^{(t)}$ to obtain $\mathcal{M}_t$
\ENDFOR
\STATE Manually annotate remaining low-confidence samples $\mathcal{D}_U^{\text{low}}$ to obtain $\mathcal{D}_L^{\text{hard}}$
\STATE Final fine-tuning on $\mathcal{D}_L^{\text{hard}}$
\end{algorithmic}
\end{algorithm}
\vspace{-0.5em}

\begin{table}[t]
\centering
\begin{tabular}{c|ccc}
\toprule
\textbf{Method} & \textbf{ACC (\%)$\uparrow$} & \textbf{NED$\uparrow$} & \textbf{FID$\downarrow$} \\
\midrule
MOSTEL~\cite{mostel}      & 37.69   & 0.557  & 49.19 \\
AnyText~\cite{anytext}     & 51.12   & 0.734  & 51.79 \\
TextCtrl~\cite{zeng2024textctrl}    & \textbf{84.67}   & \textbf{0.936}  & 43.78 \\
TextSSR~\cite{textssr}     & 79.69   & 0.924  & 45.37 \\
Flux.1 Kontext~\cite{flux}        & 12.81   & 0.460  & \textbf{41.59} \\
Qwen-Image-Edit~\cite{wu2025qwen}  & 53.16   & 0.825  & 49.91 \\
\bottomrule
\end{tabular}

\caption{Comparison of different text image generation methods on the ScenePair~\cite{zeng2024textctrl}. Correctness is measured using ACC (word accuracy) and NED (normalized edit distance), evaluated with an official text recognition model~\cite{whatwrong}. Realism is assessed using FID~\cite{heusel2017gans}, measuring the difference between feature vectors. }
\label{tab:scene_edit_comparison}
\vspace{-1.5em}
\end{table}

\begin{table*}[t]\footnotesize
\centering
\setlength{\tabcolsep}{3pt}{
\begin{tabular}{c|c|ccccccc|cccccccc}
\toprule
\multirow{2}{*}{\textbf{Training Data}} & \multirow{2}{*}{\textbf{Volume}} & \multicolumn{7}{c|}{\textbf{Common Benchmarks}}                                                                & \multicolumn{8}{c}{\textbf{Union14M-Benchmark}}                                             \\
 & &
\textit{IIIT} & \textit{SVT}  & \textit{IC13} & \textit{IC15} & \textit{SVTP} & \textit{CUTE} & \textit{AVG} &
\textit{CUR} & \textit{MLO} & \textit{ART} & \textit{CTL} & \textit{SAL} & \textit{MLW} & \textit{GEN} & \textit{AVG}                  
\\
\midrule

MJ~\cite{mj}    &   8.92M      & 88.93 & 90.73 & 95.92 & 80.29 & 86.67 & 80.21 & 87.12  & 25.60  & 24.98 & 45.67 & 30.30  & 38.98 & 15.78 & 39.38 & 31.52  \\
ST~\cite{st}    &   6.98M      & 96.23 & 92.12 & 96.97 & 84.37 & 88.99 & 92.01 & 91.78  & 65.91 & 41.56 & 58.22 & 65.47 & 73.59 & 67.11 & 62.80 & 62.10  \\
ST Family & 17.1M & 97.77 & 94.13 & 97.32 & 87.58 & 90.39 & 95.49 & 93.78  & 82.98 & 77.50 & 60.33 & 73.68 & 80.10 & 73.91 & 67.99 & 73.78  \\
MJ~+~ST & 15.9M      & 97.67 & 93.66 & 98.02 & 87.47 & 91.78 & 94.79 & 93.90  & 73.74 & 43.32 & 61.44 & 69.19 & 76.56 & 71.00 & 64.72 & 65.71  \\
SynthTIGER~\cite{yim2021synthtiger} & 10.0M  & 97.47 & 94.90 & 98.25 & 86.36 & 91.32 & 92.36 & 93.44  & 47.28 & 30.53 & 61.33 & 74.71 & 59.82 & 71.60 & 59.81 & 57.87  \\
ST-2D & 36.0M & 98.33 & 95.83 & 98.60 & \underline{88.90} & 90.85 & \underline{96.87} & 94.90  & 83.35 & 73.19 & 63.22 & 75.10 & 80.73 & 69.54 & 68.39 & 73.36  \\
UnrealText~\cite{long2020unrealtext} & 19.6M & 87.43 & 88.72 & 92.30 & 77.69 & 82.17 & 65.97 & 82.38 & 8.16 & 20.82 & 50.11 & 61.87 & 19.46 & 38.35 & 59.41 & 36.88 \\
TextSSR~\cite{textssr}  & 3.55M  &92.13	&87.17	&92.88	&78.69	&78.91	&86.81	&86.10	&46.62	&20.89	&33.89	&68.16	&57.17	&57.40	&55.07	&48.46 \\
UnionST-S   & 5.00M  & 98.20  & \underline{95.98} & 98.48 & 88.29 & 94.11 & \underline{96.87} & 95.32 & \underline{88.99} & \underline{94.45} & 73.11 & 80.62 & \underline{81.68} & 87.26 & 74.87 & 83.00 \\
UnionST-S   & 10.0M  & \underline{98.66}  & \underline{95.98} & 98.71 & 88.13 & \underline{94.88} & \underline{96.87} & \underline{95.54} & 88.13 & 93.42 & \underline{73.67} & \underline{82.80} & 79.66 & \underline{87.74} & \underline{77.62} & \underline{83.29} \\
\midrule
UnionST-P   & 5.00M  & 98.70  & 97.22 & 98.60 & 89.23 & 95.97 & 97.57 & 96.21 & 89.45 & 95.03 & 78.22 & 83.83 & 81.49 & 83.50 & 78.58 & 84.30 \\

UnionST-SP  & 10.0M & 98.60 & 97.37 & 99.07 & 89.29 & 94.88 & 97.22 & 96.07 & 90.64 & 95.69 & 75.67 & 84.98 & 80.16 & 87.99 & 78.92 & 84.86 \\

\midrule

U14M-Filter~(R)~\cite{du2024svtrv2}   & 3.22M    & 99.03 & 97.37 & 98.60 & 90.56 & 95.50 & 98.26 & 96.56 & 91.71 & 94.74 & 79.44 & 86.01 & 86.86 & 86.29 & 85.47 & 87.22 \\

UnionST-S~+~1\%R   & 5.00M~+~32.2K  & 98.93 & 97.22 & 99.07 & 90.28 & 94.88 & 98.26 & 96.44 & 92.00 & 95.76 & 81.00 & 84.47 & 86.54 & 88.71 & 82.34 & 87.26 \\

ST-2D~+~R  & 36.0M~+~3.22M & \textbf{99.47} & 98.15 & 99.18 & 91.66 & 96.43 & 99.65 & 97.42 & 94.64 & 95.84 & 84.00 & 86.52 & 90.08 & 90.53 & 86.38 & 89.71   \\

TextSSR~+~R  & 3.55M~+~3.22M &99.13	&97.84	&98.72	&91.28	&95.66	&98.61	&96.87	&92.91	&93.21	&82.33	&86.91	&88.12	&88.23	&85.97	&88.24   \\

UnionST-SP~+~R   & 10.0M~+~3.22M    & \textbf{99.47} & \textbf{98.92} & \textbf{99.30} & \textbf{92.44} & \textbf{96.90} & \textbf{99.99} & \textbf{97.84} & \textbf{95.42} & \textbf{97.22} & \textbf{86.67} & \textbf{89.60} & \textbf{91.09} & \textbf{91.99} & \textbf{87.74} & \textbf{91.39} \\

\bottomrule
\end{tabular}}
\caption{Quantitative comparison of the same STR model trained on different datasets. ST Family refers to the aggregation of ST, CurvedST~\cite{curvest}, and SynthAdd~\cite{li2019show}, and ST-2D represents the aggregation of all the above 2D-based synthetic datasets. \underline{Underlines} indicate the best results on individual synthetic datasets, while \textbf{boldface} highlights the overall best results.}
\label{tab:comp}
\end{table*}

\section{Experiments}

\subsection{Datasets and Implementation Details}

We evaluate the proposed UnionST engine and its generated data against several large-scale synthetic datasets, including MJ~\cite{mj}, ST~\cite{st}, SynthAdd~\cite{li2019show}, CurvedST~\cite{curvest}, SynthTIGER~\cite{yim2021synthtiger}, UnrealText~\cite{long2020unrealtext}, TextSSR~\cite{textssr} and their common combinations. To avoid label leakage and better simulate real scenarios, we use U14M-Filter~\cite{jiang2023revisiting, du2024svtrv2}, which removes duplicates between training and test sets, making it a more reliable data source for SEL evaluation. Evaluations are conducted on six common benchmarks (Common), including ICDAR 2013 (IC13)~\cite{icdar2013}, Street View Text (SVT)~\cite{Wang2011SVT}, IIIT5K-Words (IIIT5K)~\cite{IIIT5K}, ICDAR 2015 (IC15)~\cite{icdar2015}, Street View Text-Perspective (SVTP)~\cite{SVTP}, and CUTE80 (CUTE)~\cite{Risnumawan2014cute}), as well as the Union14M-Benchmark (U14M-Bench)~\cite{jiang2023revisiting}. It includes seven subsets: (Curve (\textit{CUR}), Multi-Oriented (\textit{MLO}), Artistic (\textit{ART}), Contextless (\textit{CTL}), Salient (\textit{SAL}), Multi-Words (\textit{MLW}), and General (\textit{GEN})). Performance is measured by the normalized accuracy, where non-vocabulary characters and spaces are removed, and all text is lowercased prior to evaluation. More details are provided in the supplementary.

\subsection{Experimental Results}

\noindent\textbf{STR Model Evaluation.} Tab.~\ref{tab:comp_str} compares models with various encoder and decoder architectures, demonstrating our model selection process. We observe that SVTRv2~\cite{du2024svtrv2}, which employs only a CTC-based decoding strategy, surpasses previous PD-based (parallel decoding) and AR-based methods. This improvement is primarily due to its advanced encoder, which adapts effectively to inputs of diverse sizes, making it suitable for challenging scenarios. Additionally, we notice that MAERec~\cite{jiang2023revisiting}, combining a ViT encoder with an AR decoder, also achieves competitive performance. Motivated by these findings, we integrate the strengths of both approaches to construct our model, SVTRv2-AR. By leveraging SOTA encoder and decoder designs, our model fully utilizes the advantages provided by difficult samples in UnionST. As a result, SVTRv2-AR achieves an average accuracy of 83.00\% on the U14M-Bench dataset using only synthetic training data.

\noindent\textbf{Generative Method Capabilities.}
We evaluate various generative models~\cite{mostel, anytext, zeng2024textctrl, textssr, flux, wu2025qwen} on the ScenePair~\cite{zeng2024textctrl} to reveal challenges in deep learning-based text synthesis methods. As shown in Tab.~\ref{tab:scene_edit_comparison}, both OCR-oriented text generation models and general generative models exhibit significant limitations in correctness when generating text in real-world scenes. Even the best-performing model achieves an editing accuracy of only 84.67\%. Directly using such synthetic data for training would inevitably propagate these errors into the recognition models, potentially degrading their performance. In contrast, although rendering-based methods may produce less visually realistic images compared to those generated by generative models, they offer a crucial advantage: absolute text correctness. This property makes rendering-based synthesis the most reliable solution for constructing training datasets. Moreover, as reported in Tab.~\ref{tab:comp}, models trained on UnionST consistently outperform those trained on large-scale deep learning-based synthetic datasets such as TextSSR~\cite{textssr}, both in direct usage and when mixed with real data. These results underscore the critical importance of text correctness in synthetic data for STR, and further validate the superiority of rendering-based approaches.

\noindent\textbf{Synthetic Data Comparison.} We first evaluate synthetic datasets independently, without seeing the real, to simulate a fully unknown scenario. As shown in Tab.~\ref{tab:comp}, although existing 2D-based synthetic datasets contain up to 36M samples, UnionST-S achieves superior performance with only 5M samples, outperforming them by 0.42\% on Common and by 9.64\% on the more challenging U14M-Bench. UnionST-S demonstrates substantial improvements across various challenging scenarios. These results suggest that increasing challenging samples effectively improves model performance. As shown in Tab.~\ref{tab:comparison_components}, prior work has partially addressed these challenges (such as Curve and Multi-Oriented), but fails to explore them thoroughly. UnionST still achieves improvements of 5.64\% and 21.26\% over ST-2D on these two subsets, approaching the real performance levels. This demonstrates that UnionST fully exploits the potential of rendering-based synthetic methods for STR. Furthermore, comparisons with the 3D-based UnrealText demonstrate that 3D rendering offers no clear advantage.

\begin{table*}[htbp]\footnotesize
\centering
\setlength{\tabcolsep}{2pt}{
\begin{tabular}{c|c|ccccccc|cccccccc}
\toprule
\multirow{2}{*}{\textbf{Training Data}} & \multirow{2}{*}{\textbf{Volume}} & \multicolumn{7}{c|}{\textbf{Common Benchmarks}}                                                                & \multicolumn{8}{c}{\textbf{Union14M-Benchmark}}                                             \\
 & &
\textit{IIIT} & \textit{SVT}  & \textit{IC13} & \textit{IC15} & \textit{SVTP} & \textit{CUTE} & \textit{AVG} &
\textit{CUR} & \textit{MLO} & \textit{ART} & \textit{CTL} & \textit{SAL} & \textit{MLW} & \textit{GEN} & \textit{AVG}                  
\\
\midrule
UnionST-SP  & 10.0M & 98.60 & 97.37 & 99.07 & 89.29 & 94.88 & 97.22 & 96.07 & 90.64 & 95.69 & 75.67 & 84.98 & 80.16 & 87.99 & 78.92 & 84.86 \\
UnionST-SP~+~$ \mathcal{D}_P^{(1)}$  & 10.0M~+~2.35M & 98.93 & 98.30 & 98.95 & 91.05 & 97.05 & \textbf{99.65} & 97.32 & 93.53 & 96.71 & 82.11 & 88.45 & 88.69 & 91.02 & 83.30 & 89.12 \\
UnionST-SP~+~$ \mathcal{D}_P^{(2)}$  & 10.0M~+~2.93M & 99.10 & 98.76 & 98.95 & 91.11 & \textbf{97.83} & \textbf{99.65} & 97.57 & 93.90 & 96.64 & 83.44 & 89.35 & 89.64 & 91.38 & 84.34 & 89.81 \\
UnionST-SP~+~$ \mathcal{D}_P^{(2)}$~+~$\mathcal{D}_L^{\text{hard}}$  & 12.93M ~+~290K & \textbf{99.30} & \textbf{99.23} & \textbf{99.42} & \textbf{92.21} & \textbf{97.83} & \textbf{99.65} & \textbf{97.94} & \textbf{95.05} & \textbf{96.79} & \textbf{85.56} & \textbf{91.14} & \textbf{90.65} & \textbf{92.23} & \textbf{87.17} & \textbf{91.23} \\
\midrule
UnionST-SP~+~$ \mathcal{D}_P^{(3)}$  & 10.0M~+~3.02M & 99.03 & 98.45 & 99.07 & 91.17 & 97.67 & \textbf{99.65} & 97.51 & 93.98 & 96.64 & 83.67 & 88.83 & 89.51 & 91.50 & 84.50 & 89.81 \\

\bottomrule
\end{tabular}}
\caption{Results for ISR. ``$P$'' denotes pseudo labels and ``$L$'' refers to human annotations. We adopt the definitions as specified in Sec.~\ref{sec:3-3}.}
\label{tab:isr}
\end{table*}

\begin{table*}[htb]\footnotesize
  \centering
  \setlength{\tabcolsep}{3pt}
  \begin{tabular}{c|ccccc|ccc|cccccccc}
    \toprule
    \multirow{2}{*}{\textbf{Volume}} & \multicolumn{2}{c}{\textbf{Layout}} & \multirow{2}{*}{\textbf{Corpus}} & \multirow{2}{*}{\textbf{Font}} & \multirow{2}{*}{\textbf{DTAug}} & \multicolumn{3}{c|}{\textbf{Common Benchmarks}} & \multicolumn{8}{c}{\textbf{Union14M-Benchmark}} \\
    & Curve & Multi-Oriented &  & & & Regular & Irregular  & Average & \textit{CUR} & \textit{MLO} & \textit{ART} & \textit{CTL} & \textit{SAL} & \textit{MLW} & \textit{GEN} & \textit{AVG} \\
\midrule

\multirow{10}{*}{0.5M} &             &             &             &             &     &   93.99	& 82.62 & 88.30   & 19.83 & 30.39 & 51.78 & 56.99 & 28.30 & 59.59 & 48.96 & 42.26 \\
 & \ding{51}   &             &             &             &      & 94.35	& 84.49  &  89.42   & 46.70 & 33.31 & 48.78 & 60.21 & 32.03 & 61.41 & 49.00 & 47.35 \\
 & \ding{51}   & \ding{51}   &             &             &        & 94.68	& 86.42  & 90.55  & 76.55 & 87.29 & 52.11 & 59.69 & 57.11 & 61.17 & 52.82 & 63.82 \\
 & \ding{51}   & \ding{51}   & \ding{51}   &             &      & 94.66	& 86.54   & 90.60   & 75.93 & \textbf{88.31} & 61.33 & 70.73 & 64.31 & \textbf{79.73} & 53.86 & 70.60 \\
 & \ding{51}   & \ding{51}   & \ding{51}   & \ding{51}   &    &  94.56	& 86.09   & 90.32   & 75.85 & 87.80 & 58.44 & 68.68 & 62.22 & 78.40 & 53.69 & 69.30 \\
 & \ding{51}   & \ding{51}   & \ding{51}   & \ding{51}   & \ding{51} & 95.46	& 87.68 & 91.57 & 75.76 & 87.80 & 62.33 & 70.60 & 63.74 & 78.88 & \textbf{65.27} & \textbf{72.06} \\
 & \ding{51}   & \ding{51}   & \ding{51}   &             & \ding{51} & 95.07	& 87.38 & 91.22  & 76.42 & 87.80 & 59.00 & 70.22 & 64.88 & 78.64 & 64.40 & 71.62 \\
 & \ding{51}   & \ding{51}   &             & \ding{51}   & \ding{51} & 95.41	& \textbf{88.52} & \textbf{91.96} & \textbf{77.00} & 86.92 & 58.89 & 56.10 & 61.34 & 59.83 & 63.10 & 66.17 \\
 & \ding{51}   &             & \ding{51}   & \ding{51}   & \ding{51} & \textbf{95.98}	& 87.11 & 91.54  & 50.00 & 34.26 & 61.67 & \textbf{71.12} & 41.25 & 77.55 & 61.89 & 56.82 \\
 &             & \ding{51}   & \ding{51}   & \ding{51}   & \ding{51} & 95.52	& 87.26 & 91.39  & 62.32 & 85.98 & \textbf{63.44} & 69.45 & \textbf{66.01} & 77.06 & 65.21 & 69.93 \\
\midrule
\multirow{2}{*}{5.0M} & \ding{51}   & \ding{51}   & \ding{51}   &             & \ding{51} & 97.00	& 92.02 & 94.51  & 85.82 & 92.18 & 62.56 & 77.66 & 78.52 & 83.37 & 74.66 & 79.25 \\
& \ding{51}   & \ding{51}   & \ding{51}   & \ding{51}   & \ding{51} & \textbf{97.55}	& \textbf{93.09} & \textbf{95.32}  & \textbf{88.99} & \textbf{94.45} & \textbf{73.11} & \textbf{80.62} & \textbf{81.68} & \textbf{87.26} & \textbf{74.87} & \textbf{83.00} \\

\bottomrule
    \end{tabular}
    \caption{For control experiments in which Curve, Multi-Oriented, Corpus, and Font are not selected, we apply the following settings: $a$ is always set to 0 to simulate straight placement, $\phi$ is set to 0 to simulate the absence of multi-orientation, the `MJ+ST' corpus is used as the baseline, and the filtered Google Fonts collection is utilized. When the data volume is 0.5M, the model is trained for 100 epochs; for a volume of 5.0M, training is conducted for 80 epochs. All other training configurations are consistent with those described previously. ``Regular” denotes the average on \textit{IIIT}, \textit{SVT}, and \textit{IC13} datasets,``Irregular” represents the average on \textit{IC15}, \textit{SVTP}, and \textit{CUTE} datasets.}
    \label{tab:ablation_studies}
\end{table*}

\noindent\textbf{The Incorporation of Pseudo-Corpus.} With the same scale, UnionST-P increases average accuracy by 1.30\% over UnionST-S on U14M-Bench. Furthermore, we observe that even when scaling UnionST-S to 10M samples, its performance exhibits only marginal improvement and still lags behind the UnionST-SP by a noticeable margin. These results suggest that 5M-scale is sufficient and highlight the importance of considering a more realistic distribution.

\noindent\textbf{Synthetic Data Potential.} We further investigate the necessity of UnionST in large-scale real data scenarios. Pre-training with UnionST-SP increases the average accuracy by 4.17\% on U14M-Bench, outperforming the improvement achieved by ST-2D by 1.68\%. Notably, it raises the average accuracy on the U14M-Bench to 91.39\%. To our knowledge, this is the first time that STR performance has exceeded 90\% on this benchmark. This also demonstrates the significance of synthetic data in the STR field. Even for English, the current real data remains insufficient, and further support from synthetic data is needed. 

\noindent\textbf{Limited Real Data Access.} We simulate low-resource scenarios by fine-tuning with a small and randoml subset of labeled real data. When the UnionST-S trained model is fine-tuned with only 1\% of it, the model achieves an average accuracy of 87.26\% on U14M-Bench, which is comparable to the 87.22\% obtained by using the full real dataset.

\noindent\textbf{ISR Results.} As shown in Tab.~\ref{tab:isr}, after the first iteration, the accuracy reaches 89.12\% on U14M-Bench. The second iteration further improves performance to 89.81\%. However, the third iteration shows no further gains, due to the trade-off between error and the increase in labeled samples. This balance is further illustrated in the top visualization of Fig.~\ref{fig:threshold_selection}, where the number of correct samples continues to grow, but the overall accuracy declines. Leveraging the results from the second iteration, we simulate minimal manual annotation by incorporating 290K manually annotated hard samples, which increases accuracy to 91.23\%. This result is only 0.16\% lower than the accuracy achieved using both synthetic and fully annotated real data, while reducing annotation costs from 3.22M to 290K, a 91\% reduction.

\begin{table*}[t]\footnotesize
\centering
\setlength{\tabcolsep}{3pt}{
\begin{tabular}{c|c|ccccccc|cccccccc}
\toprule
\multirow{2}{*}{\textbf{Training Data}} & \multirow{2}{*}{\textbf{Volume}} & \multicolumn{7}{c|}{\textbf{Common Benchmarks}}                                                                & \multicolumn{8}{c}{\textbf{Union14M-Benchmark}}                                             \\
 & &
\textit{IIIT} & \textit{SVT}  & \textit{IC13} & \textit{IC15} & \textit{SVTP} & \textit{CUTE} & \textit{AVG} &
\textit{CUR} & \textit{MLO} & \textit{ART} & \textit{CTL} & \textit{SAL} & \textit{MLW} & \textit{GEN} & \textit{AVG}                  
\\
\midrule
MJ+ST~+~$ \mathcal{D}_P'^{(1)}$  & 15.9M~+~2.41M & 98.63 & 95.67 & 98.48 & 89.78 & 93.49 & 96.18 & 95.37 & 83.92 & 47.99 & 69.22 & 77.02 & 82.69 & 81.19 & 71.08 & 73.30 \\
ST-2D~+~$ \mathcal{D}_P''^{(1)}$  & 36.0M~+~2.36M & 98.93 & 97.68 & 98.48 & 90.72 & 93.02 & 97.92 & 96.13 & 89.82 & 83.35 & 71.22 & 80.49 & 86.48 & 76.33 & 75.87 & 80.51 \\
UnionST-S~+~$ \mathcal{D}_P'''^{(1)}$  & 5.00M~+~2.24M & 98.80 & 97.84 & 98.72 & 90.61 & 96.59 & 98.61 & 96.86 & 92.79 & 96.42 & 79.44 & 87.16 & \textbf{88.95} & 90.66 & 81.23 & 88.09 \\
UnionST-SP~+~$ \mathcal{D}_P^{(1)}$  & 10.0M~+~2.35M & \textbf{98.93} & \textbf{98.30} & \textbf{98.95} & \textbf{91.05} & \textbf{97.05} & \textbf{99.65} & \textbf{97.32} & \textbf{93.53} & \textbf{96.71} & \textbf{82.11} & \textbf{88.45} & 88.69 & \textbf{91.02} & \textbf{83.30} & \textbf{89.12} \\

\bottomrule
\end{tabular}}
\caption{Experiments on base data for ISR. The superscripts of different ``$\mathcal{D}_P^{(1)}$'' denote different base data. At the same threshold and using the same real dataset, they correspond to the pseudo-label data obtained during the first iteration.}
\label{tab:comp_2}
\end{table*}

\subsection{Ablation Studies}

Tab.~\ref{tab:ablation_studies} presents the performance gains achieved by progressively integrating components into the UnionST engine. We incrementally introduce components and conduct single-component ablation experiments after the full system is assembled. The curved placement has a pronounced effect on the Curve subset.  Similarly, the multi-oriented variation significantly impacts the Curve, Multi-Oriented, and Salient subsets.  Corpus augmentation primarily benefits the Contextless and Multi-Words subsets, but may cause a slight performance drop on Common. This also indicates that most examples in Common are common words, and evaluating only on them can easily lead to overfitting. DTAug yields notable improvements on the General subset.  Font diversity has little effect at small data scales.  However, reducing font variety at 5M-scale leads to a marked drop in Artistic subset.  This series of experiments also clarify what's wrong with previous rendering-based methods.  They fail to fully consider the need for diverse and challenging modeling of each component, and do not combine these challenges to simulate various texts in the wild.

\begin{figure}[t]
\centering
\includegraphics[width=0.4\textwidth]{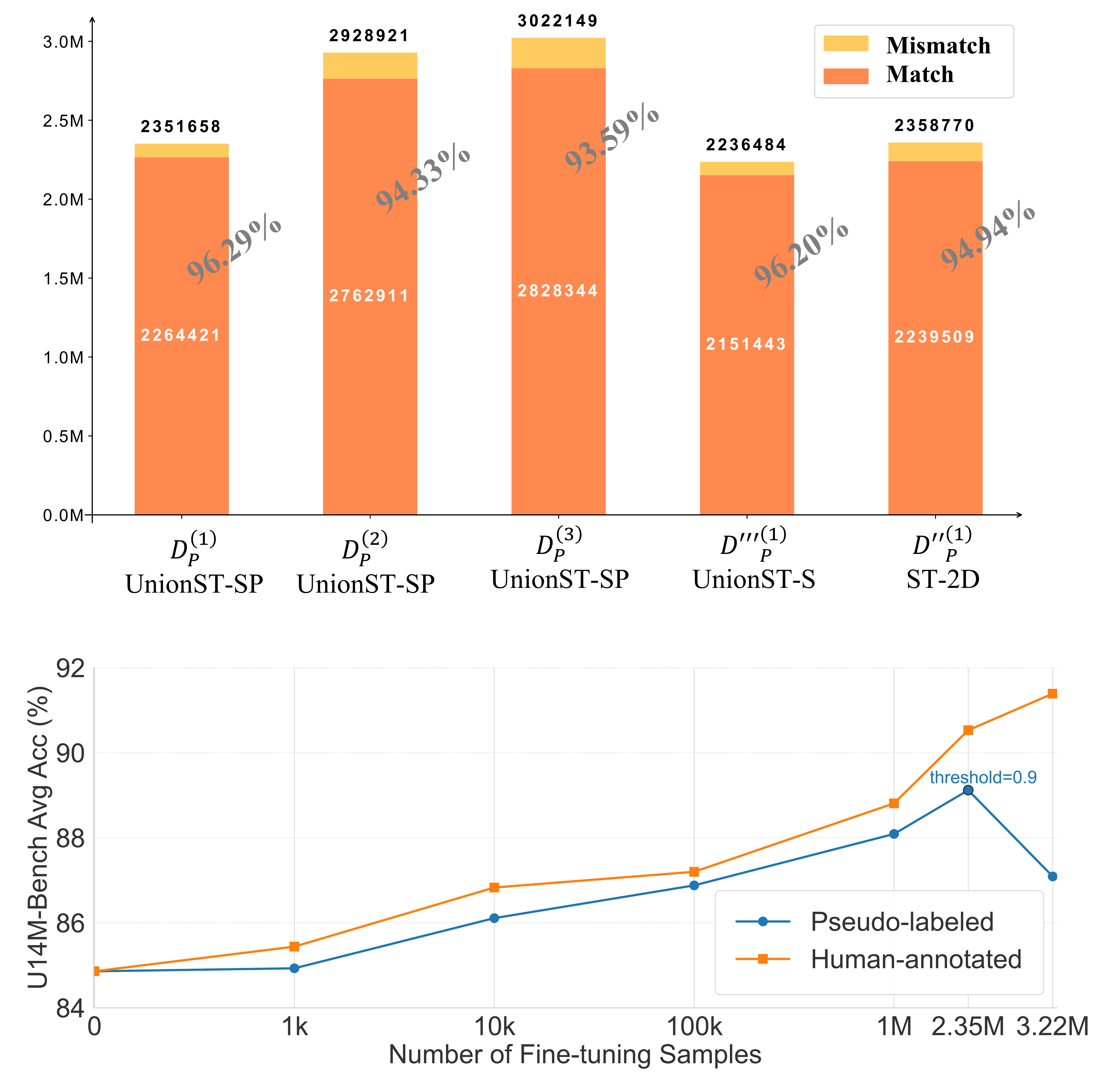} 
\caption{\textbf{Top}: Comparison of data filtered by different data or stages for high confidence with the original human-annotated data match. \textbf{Bottom}: Comparison of accuracy trends in pseudo-labeled and human-annotated samples.}  
\label{fig:threshold_selection}
\vspace{-1em}
\end{figure}

To evaluate UnionST-SP as a foundational data for ISR, we replace it with other synthetic datasets (see Tab.~\ref{tab:comp_2}). Employing the MJ+ST or ST-2D datasets leads to improvements, reaching 73.30\% and 80.51\% on U14M-Bench, respectively. However, their limited diversity and challenge prevents them from achieving near-human annotation performance through iterative refinement. Similarly, using UnionST-S for the initial iteration yields substantial improvements, yet still lags 1.03\% behind UnionST-SP. This further demonstrates the indispensability of pseudo-corpus augmentation in the SEL framework. The matching between pseudo labels and original labels is shown in the top of Fig.~\ref{fig:threshold_selection}. It can be seen that under the same threshold (0.9), UnionST-SP produces more and more accurate labels.

The bottom of Fig.~\ref{fig:threshold_selection} illustrates the optimal threshold selection. We select pseudo-labels from models trained on synthetic data and fine-tune using descending confidence thresholds: Top-1K, Top-10K, Top-100K, Top-1M, Top-2.35M (threshold 0.9), and Top-3.22M (threshold 0). Fine-tuning with pseudo-labels initially improves performance, but accuracy declines as the confidence threshold decreases, indicating an optimal threshold near 0.9 (between Top-1M and Top-3.22M samples). In contrast, performance with human-annotated data consistently improves as data volume increases. Although pseudo-label performance remains below that of human-annotated data, ISR progressively narrows this gap, reducing reliance on manual labeling.

\section{Conclusion}
In this work, we observe that existing synthetic datasets often perform poorly on challenging real-world benchmarks. Upon analyzing prior efforts, we find that previous methods have struggled to adequately simulate the diversity of corpora, fonts, and text placements. To address these limitations, we present \textbf{UnionST}, a strong rendering-based data engine designed to significantly enhance the generation of challenging samples. Based on it we have constructed two datasets, \textbf{UnionST-S} and \textbf{UnionST-P}, and introduced an SEL framework for selective labeling. Experimental results demonstrate that our synthetic datasets consistently outperform conventional ones, even when using fewer samples. Moreover, by using our SEL framework, we achieve performance comparable to models trained on large-scale real-world annotations, while substantially reducing the cost of manual labeling. This approach not only narrows the domain gap between synthetic and real data, but also provides a cost-effective solution for scenarios with limited annotation resources. In future, we plan to pick out commonly used document and handwriting fonts from the general font set, evaluate our method under these settings, and extend it to domains such as document OCR \cite{du2025unirec} and handwriting recognition \cite{garrido2025handwritten}. Furthermore, we are interested in developing a multilingual version to support low-resource STR.

\noindent\textbf{Acknowledgment}
This work was supported by the National Natural Science Foundation of China under Grants 625B2057 and 32341012.

{
    \small
    \bibliographystyle{ieeenat_fullname}
    \bibliography{main}
}

\clearpage
\setcounter{page}{1}
\maketitlesupplementary

\begin{figure}[t]
\centering
\includegraphics[width=0.48\textwidth]{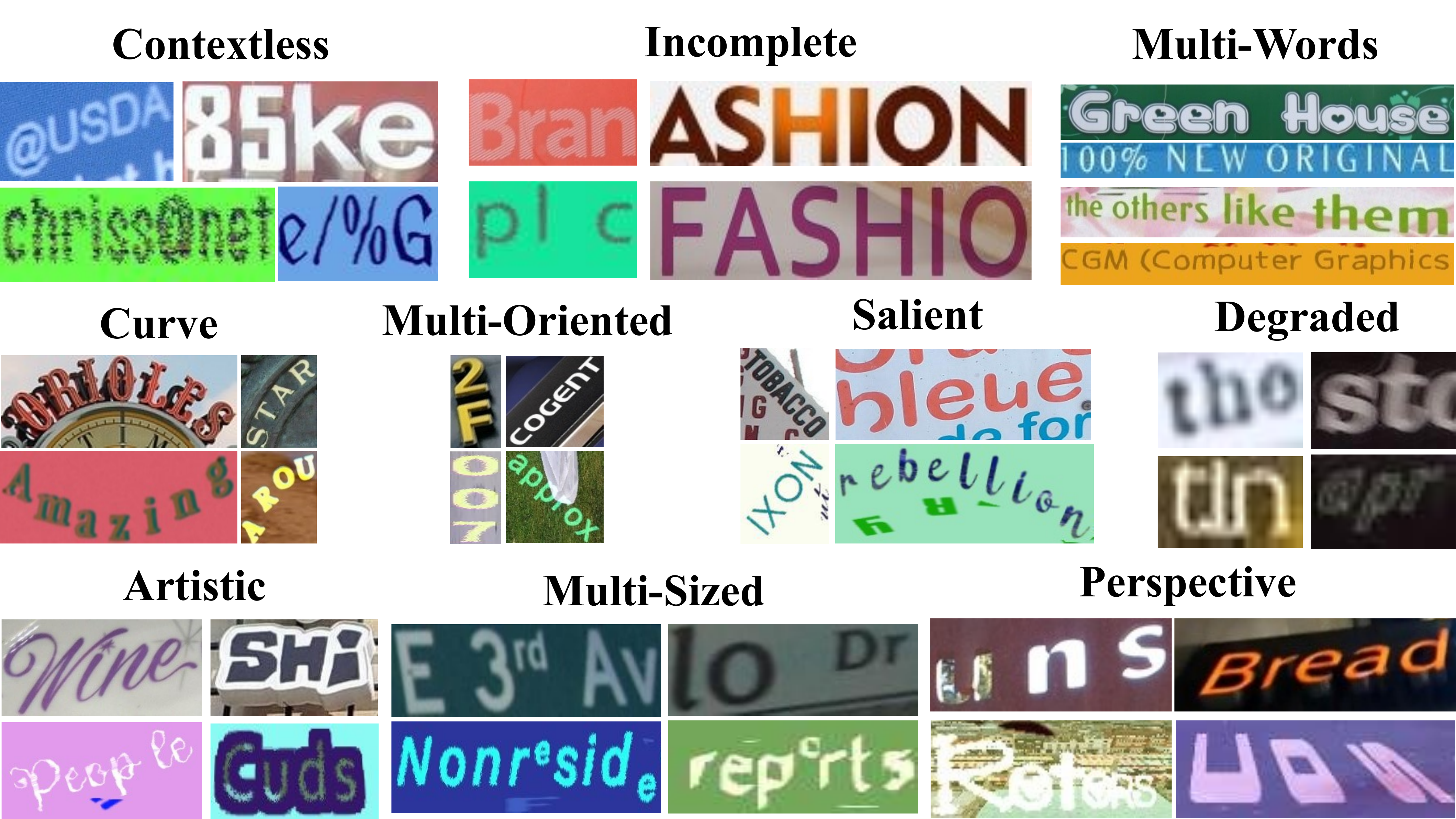} 
\caption{Real text samples (the top line in each subset) and data generated by UnionST (the bottom line) grouped by subsets according to their challenges. Beyond the subsets identified in Union14M~\cite{jiang2023revisiting}, we introduce three additional ones: \textit{Multi-Sized} (words with varying sizes, including subscripts and superscripts), \textit{Perspective} (variations in viewpoint), and \textit{Degraded} (blur or low resolution caused by camera shake or small text size). }  
\label{fig:data_view}
\end{figure}

\section{UnionST Details}

\subsection{Corpus} The details of the previously mentioned corpus construction are as follows:

\begin{itemize}
\item \textbf{Common:} We augment the MJ~\cite{mj}\&ST~\cite{st} corpora, applying case transformations (original, lowercase, uppercase, capitalized) to obtain 264K and 13M samples, respectively. 
\item \textbf{Contextless}: Random character sequences (length 2--25, 10K per length, totaling 240K) are generated from a character set consisting of 94 commonly used alphabetic and symbolic characters.
\item \textbf{Incomplete}: We create incomplete words by randomly removing initial, terminal, or internal characters from words in the MJ corpus, yielding 264K samples.
\item \textbf{Multi-Words}: We collect 400K common phrases and concatenated multi-word expressions. Additionally, we extract substrings of varying lengths (1--25, 120K per length, totaling 3M) from the ST newspaper corpus.
\end{itemize}

\begin{figure}[t]
\centering
\includegraphics[width=0.48\textwidth]{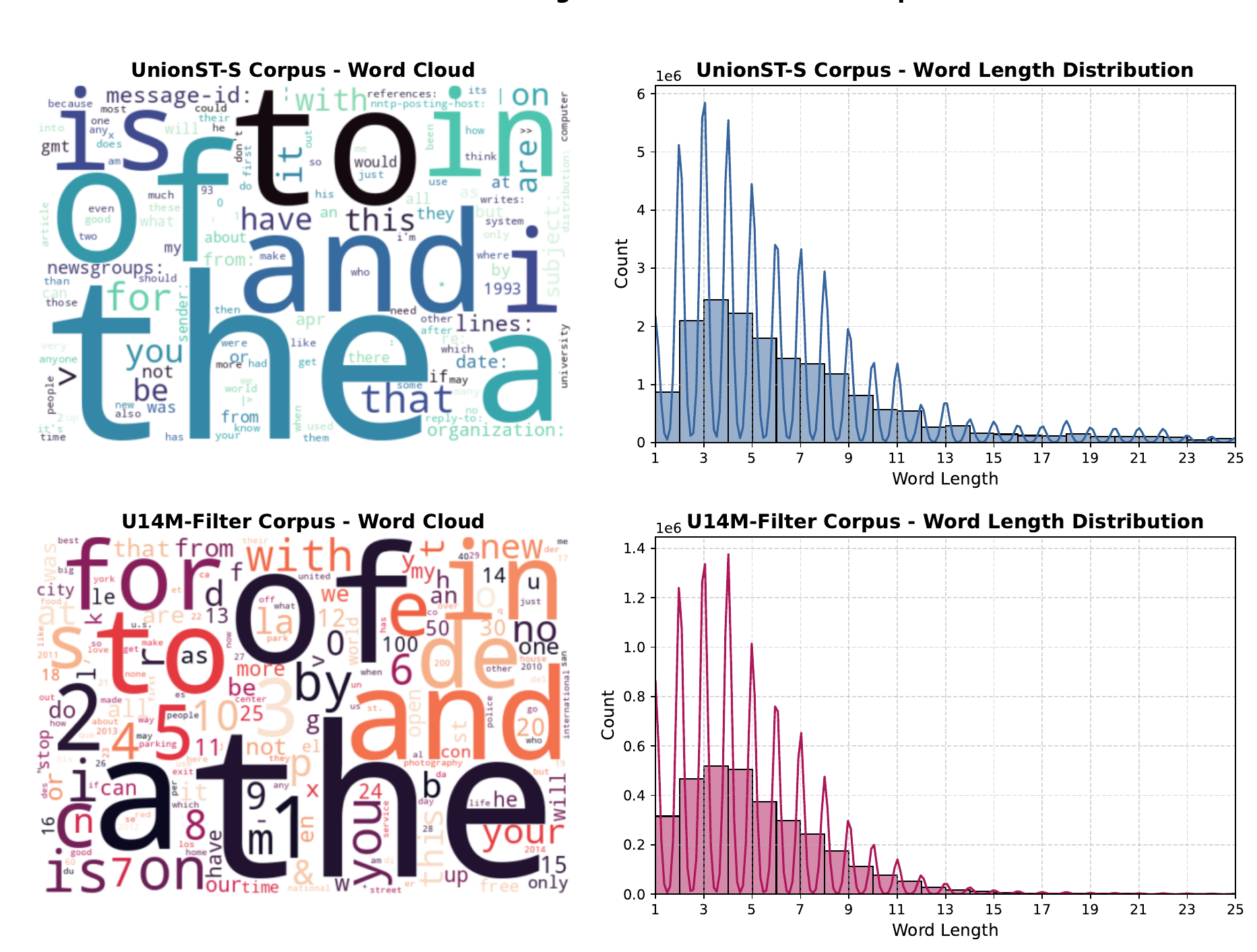} 
\caption{Word clouds (left) and word length distributions (right) for the synthetic corpus (top) and the real (bottom).} 
\label{fig:word_length}
\end{figure}

\begin{figure}[t]
\centering
\includegraphics[width=0.48\textwidth]{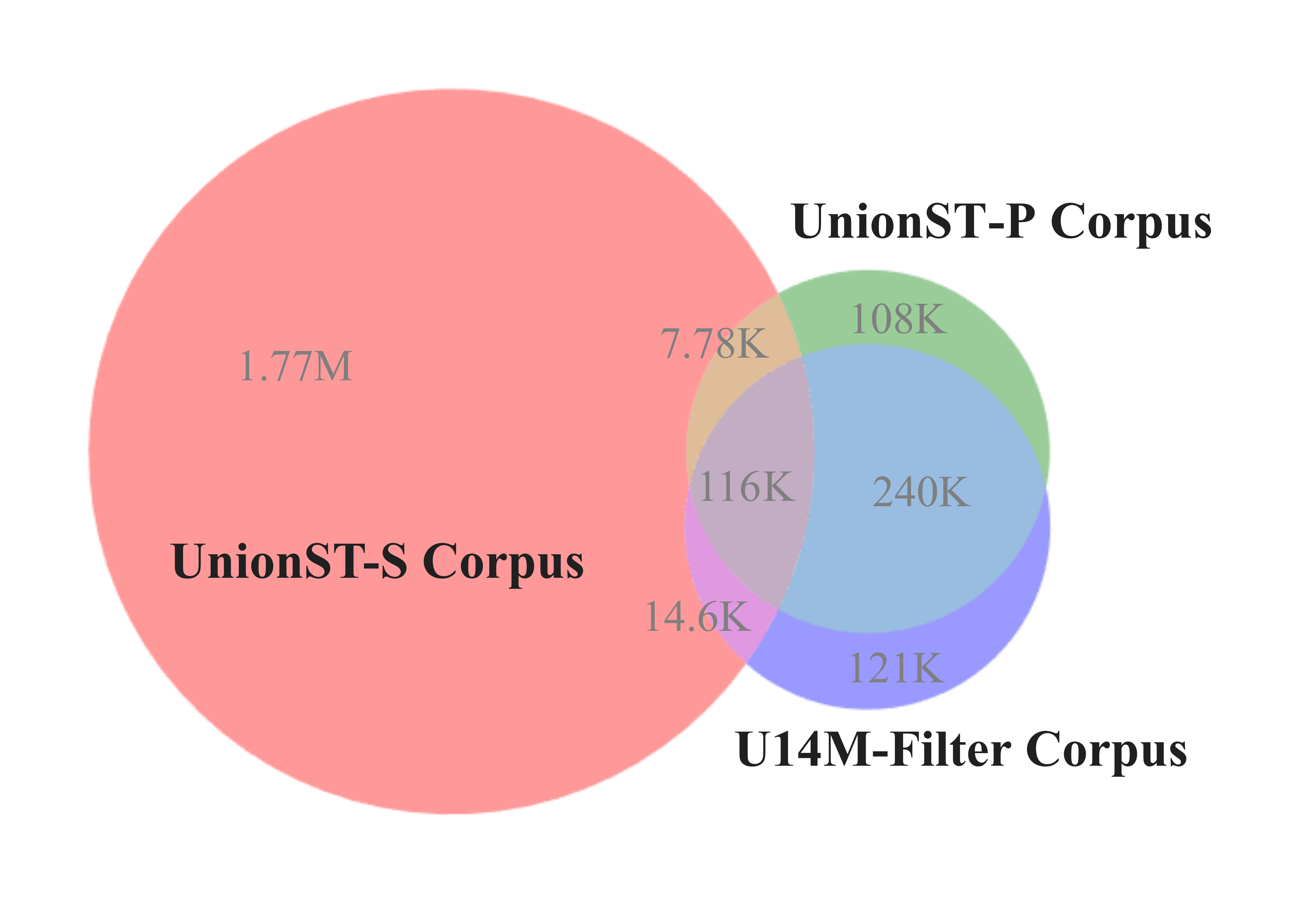} 
\caption{The Venn diagram comparing words from three different corpora reveals a substantial disparity between the synthesized corpus and the real corpus.}  
\label{fig:corpus_venn}
\end{figure}

\begin{figure*}[t]
\centering
\includegraphics[width=0.95\textwidth]{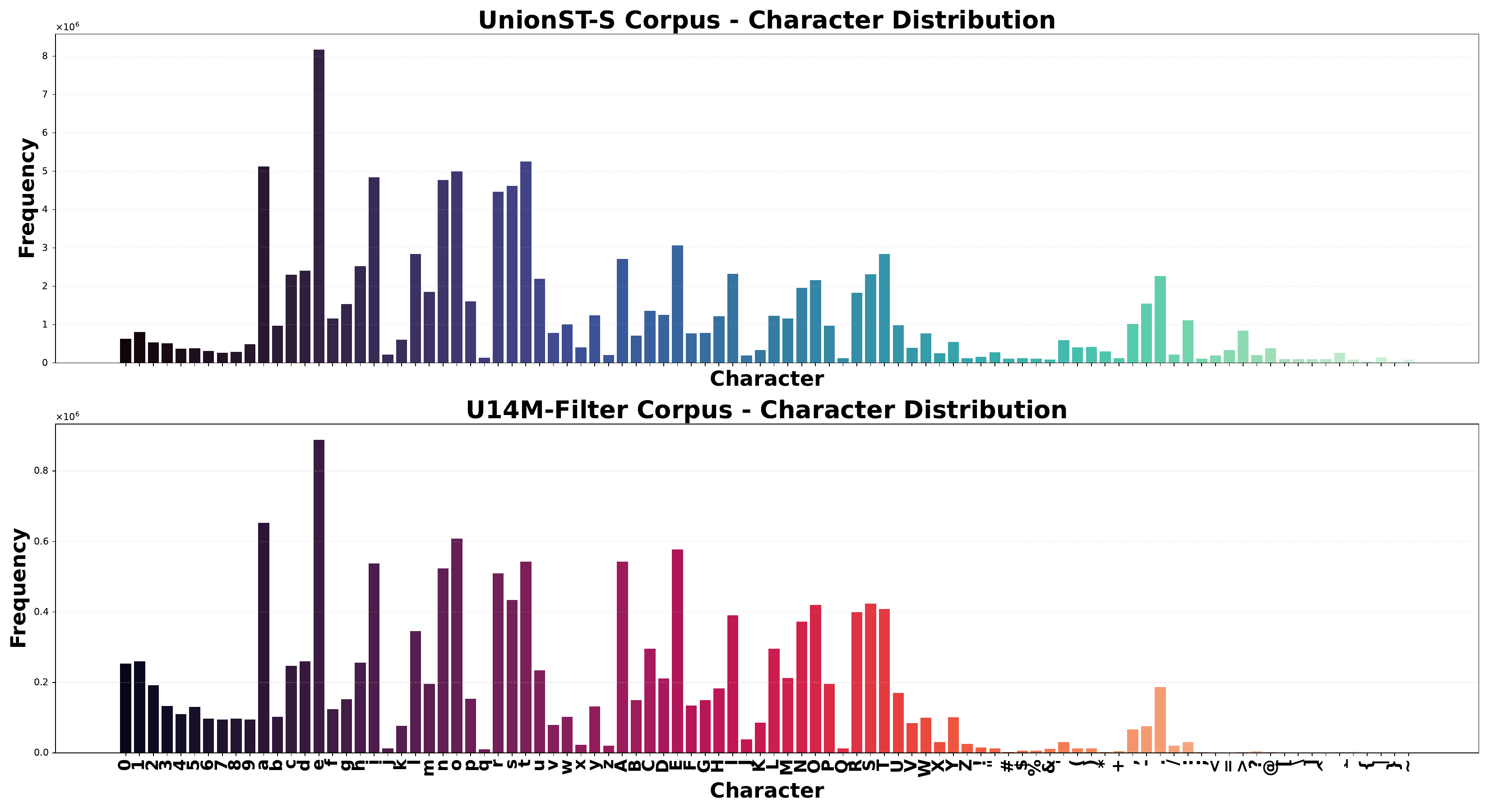} 
\caption{Character distribution of the UnionST-S Corpus (top) and the U14M-Filter Corpus (bottom).} 
\label{fig:char_dis}
\end{figure*}

Fig.~\ref{fig:word_length} shows that the UnionST-S corpus demonstrates a word length distribution for short texts that closely aligns with the real dataset, while also providing a higher proportion of longer texts. Fig.~\ref{fig:char_dis} further shows that the UnionST-S corpus places more emphasis on certain rare or special characters. And Fig.~\ref{fig:corpus_venn} illustrates the word distribution among different corpora in UnionST. It can be seen that the pseudo-labeling approach enables us to obtain results at scale that closely match those of real corpora, leaving only 121K words in UnionST-SP uncovered. At the same time, this approach provides a much larger corpus overall.

\subsection{Font} 
We curate 113,788 font files from publicly available open-source repositories. For ablation studies, we use a filtered subset of 3,092 fonts from the Google Fonts collection for comparison. During the collection process, we pay special attention to the license terms, ensuring that all fonts are clearly labeled as free for commercial use at their source websites or repositories, and comply with the corresponding open-source licenses (such as SIL Open Font License, Apache License, etc.).

\subsection{Other Resources} 
Other details not mentioned are basically consistent with ST~\cite{st} and SynthTIGER~\cite{yim2021synthtiger}, such as the color mapping table, background images, and the filtering mechanism after generating images, etc.

\subsection{Other Algorithm Improvements} 
Real images often contain distracting non-target text as mid-ground elements. Following SynthTIGER's strategy~\cite{yim2021synthtiger}, we also add non-salient text as background clutter to focus on the main salient text. However, we find that SynthTIGER's implementation has a problem: when handling foreground and mid-ground text, it often fails to account for the differences in their respective sizes. This results in simulated mid-ground text that lacks diversity. As shown in Alg.~\ref{alg_mid}, the \underline{underlined} part indicates what we add.

\begin{algorithm}[t]
\caption{Improved Mid-ground Text Blending}
\label{alg:midground_blending}
\begin{algorithmic}[1]
\REQUIRE Foreground text $F$, Mid-ground text $M$, Background image $I_{bg}$
\ENSURE Composite image $I$
\STATE \underline{$\text{placement}_M,~ \text{placement}_F  \leftarrow \text{Compute}(F, M, I_{bg})$
}
\STATE \underline{$I_{crop} \leftarrow \text{Crop}(I_{bg})$}
\STATE $I_{mid} \leftarrow \text{Blend}(I_{crop}, M, \text{placement}_M)$
\STATE $I_{mid} \leftarrow \text{EraseOverlap}(I_{mid}, F, \text{placement}_F)$
\STATE $I \leftarrow \text{Blend}(I_{mid}, F, \text{placement}_F)$
\RETURN $I$
\end{algorithmic}
\label{alg_mid}
\end{algorithm}

\subsection{Dataset Properties}
\begin{figure}[ht]
 \centering
 \includegraphics[width=0.9\linewidth]{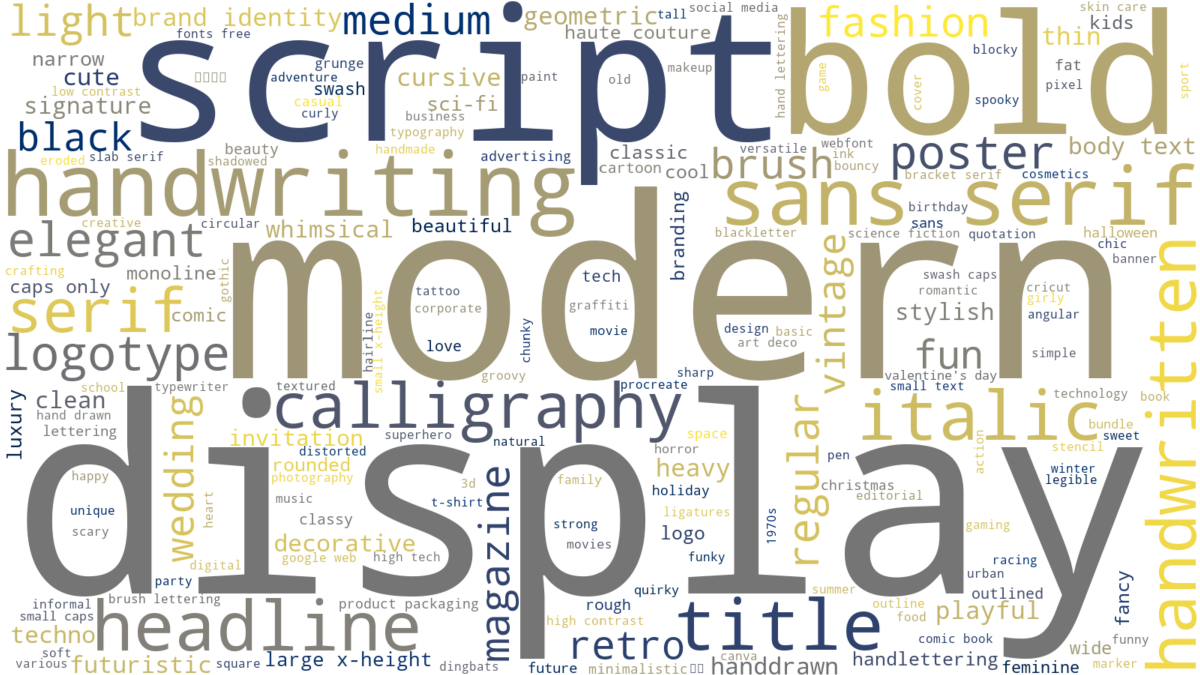}
 \caption{Word cloud of font tags uesd in UnionST.}
 \label{fig:font_tags}
\end{figure}

Regarding image types, we store all images in the JPEG format and write them into the lmdb files that are commonly used in STR. Fig.~\ref{fig:font_tags} shows the distribution of tags, such as modern, display, handwriting, and script, demonstrating the comprehensiveness and diversity of our font coverage.

\subsection{Other Variants}

If real labels are available, we can naturally use them as the labels for the synthetic data. Therefore, we use the text labels from U14M-Filter~\cite{jiang2023revisiting, du2024svtrv2} as the corpus to construct UnionST-R (also 5M). This allows us to evaluate the performance of UnionST-P.

\subsection{Runtime and Cost Efficiency}
For UnionST data synthesis, we employ a server equipped with an Intel Xeon Platinum 8255C CPU (96 cores, 375~GB RAM).    With 16 worker processes, generating 1 million samples requires 49,040 seconds, corresponding to a throughput of approximately 20.39 samples per second.    Compared to deep generative methods, which often rely heavily on GPU resources, UnionST offers significant advantages in both cost and computational efficiency.    The CPU cost for UnionST is \$0.2 per hour, so generating 5 million samples incurs a total cost of only \$13.62.    For reference, TextSSR reports that a single RTX-3090 GPU can generate 6.15K images per card-hour, meaning that producing 5 million samples requires 813 card-hours and costs \$325.2 at \$0.4 per card-hour.   Closed-source Nano Banana charges \$0.039 per image, resulting in a \$195,000 budget for producing 5 million samples.  Additionally, the quoted price for manual annotation is \$0.06 per image, meaning annotating 5 million images would cost \$300,000. Our SEL framework reduces annotation costs to \$27,000 and can provide pseudo-labels to accelerate annotation efficiency.  Overall, UnionST achieves far lower costs, demonstrating clear resource-efficient characteristics and suitability for large-scale synthetic data generation.

\begin{table*}[t]\footnotesize
\centering
\setlength{\tabcolsep}{2pt}{
\begin{tabular}{c|c|c|ccccccc|cccccccc}
\toprule
\multirow{2}{*}{\textbf{Type}} &\multirow{2}{*}{\textbf{Model}} & \multirow{2}{*}{\textbf{Training Data}} & \multicolumn{7}{c|}{\textbf{Common Benchmarks}}                                                                & \multicolumn{8}{c}{\textbf{Union14M-Benchmark}}                                             \\
 & & &
\textit{IIIT} & \textit{SVT}  & \textit{IC13} & \textit{IC15} & \textit{SVTP} & \textit{CUTE} & \textit{AVG} &
\textit{CUR} & \textit{MLO} & \textit{ART} & \textit{CTL} & \textit{SAL} & \textit{MLW} & \textit{GEN} & \textit{AVG}                  
\\

\midrule
\multirow{12}{*}{ \textbf{CTC} } & \multirow{3}{*}{ CRNN~\cite{shi2017crnn} } &
U14M-Filter & 95.80 &91.80 &94.60 &84.90 &83.10 &91.00 &90.21 &48.10 &13.00 &51.20 &62.30 &41.40 &60.40 &68.20 &49.24 \\
& &  UnionST-S  &92.90 &85.63 &91.83 &75.37 &76.12 &82.29 &84.02 &33.59 &64.21 &36.33 &49.29 &16.49 &65.78 &48.98 &44.95 \\
& & $\Delta$ &
\textcolor{red}{-2.90} & \textcolor{red}{-6.17} & \textcolor{red}{-2.77} & \textcolor{red}{-9.53} & \textcolor{red}{-6.98} & \textcolor{red}{-8.71} & \textcolor{red}{-6.19} & \textcolor{red}{-14.51} & \textcolor{blue}{51.21} & \textcolor{red}{-14.87} & \textcolor{red}{-13.01} & \textcolor{red}{-24.91} & \textcolor{blue}{5.38} & \textcolor{red}{-19.22} & \textcolor{red}{-4.29} \\

\cline{2-18}
& \multirow{3}{*}{ ViT-CTC } &
U14M-Filter & 97.20 &	94.13 &	95.45	 & 86.03 &	87.75 &	92.01 &	92.10	& 70.03	& 80.50 &	58.00	 & 72.53	& 68.04 & 	73.06 &	77.56	& 71.39 \\
&  & UnionST-S &   96.33 &	93.35 &	97.20 &	83.49 &	88.22 &	92.71 &	91.88 &	68.47	& 84.59 & 	58.00 &	71.12	& 52.24 &	79.61 &	65.71 &	68.53      \\
& & $\Delta$ 
& \textcolor{red}{-0.87} 
& \textcolor{red}{-0.78} 
& \textcolor{blue}{1.75} 
& \textcolor{red}{-2.54} 
& \textcolor{blue}{0.47} 
& \textcolor{blue}{0.70} 
& \textcolor{red}{-0.22} 
& \textcolor{red}{-1.56} 
& \textcolor{blue}{4.09} 
& 0.00 
& \textcolor{red}{-1.41} 
& \textcolor{red}{-15.80} 
& \textcolor{blue}{6.55} 
& \textcolor{red}{-11.85} 
& \textcolor{red}{-2.86} 
\\

\cline{2-18}
& \multirow{3}{*}{ SVTR~\cite{duijcai2022svtr} } &
U14M-Filter & 98.00 &97.10 &97.30 &88.60 &90.70 &95.80 &94.58 &76.20 &44.50 &67.80 &78.70 &75.20 &77.90 &77.80 &71.17 \\
&  & UnionST-S  &97.50 &94.28 &97.78 &85.64 &90.08 &95.49 &93.46 &74.73 &86.56 &62.44 &74.45 &62.67 &81.80 &68.67 &73.05 \\
& & $\Delta$ &
\textcolor{red}{-0.50} & \textcolor{red}{-2.82} & \textcolor{blue}{0.48} & \textcolor{red}{-2.96} & \textcolor{red}{-0.62} & \textcolor{red}{-0.31} & \textcolor{red}{-1.12} & \textcolor{red}{-1.47} & \textcolor{blue}{42.06} & \textcolor{red}{-5.36} & \textcolor{red}{-4.25} & \textcolor{red}{-12.53} & \textcolor{blue}{3.90} & \textcolor{red}{-9.13} & \textcolor{blue}{1.88} \\

\cline{2-18}
& \multirow{3}{*}{ SVTRv2~\cite{du2024svtrv2} }  & 
U14M-Filter &99.20 &98.00 &98.70 &91.10 &93.50 &99.00 &96.57 &90.60 &89.00 &79.30 &86.10 &86.20 &86.70 &85.10 &86.14 \\
& &  UnionST-S  & 98.20 &	95.98 &	98.60 &	87.85 &	94.26 &	97.22 & 95.35  & 83.39	& 91.31	& 67.11	& 77.41	& 70.31 &	83.25	& 71.10 & 77.70 \\
& & $\Delta$ &
\textcolor{red}{-1.00} & \textcolor{red}{-2.02} & \textcolor{red}{-0.10} & \textcolor{red}{-3.25} & \textcolor{blue}{0.76} & \textcolor{red}{-1.78} & \textcolor{red}{-1.22} & \textcolor{red}{-7.21} & \textcolor{blue}{2.31} & \textcolor{red}{-12.19} & \textcolor{red}{-8.69} & \textcolor{red}{-15.89} & \textcolor{blue}{3.45} & \textcolor{red}{-14.00} & \textcolor{red}{-8.44} \\

\midrule

\multirow{12}{*}{ \textbf{PD} } & \multirow{3}{*}{ ABINet~\cite{abinet} } &
U14M-Filter & 98.50 &98.10 &97.70 &90.10 &94.10 &96.50 &95.83 &80.40 &69.00 &71.70 &74.70 &77.60 &76.80 &79.80 &75.72 \\
& & UnionST-S  &97.00 &93.35 &96.97 &85.04 &90.39 &94.10 &92.81 &73.04 &85.10 &58.44 &64.31 &59.25 &77.31 &63.98 &68.78 \\
& & $\Delta$ &
\textcolor{red}{-1.50} & \textcolor{red}{-4.75} & \textcolor{red}{-0.73} & \textcolor{red}{-5.06} & \textcolor{red}{-3.71} & \textcolor{red}{-2.40} & \textcolor{red}{-3.02} & \textcolor{red}{-7.36} & \textcolor{blue}{16.10} & \textcolor{red}{-13.26} & \textcolor{red}{-10.39} & \textcolor{red}{-18.35} & \textcolor{blue}{0.51} & \textcolor{red}{-15.82} & \textcolor{red}{-6.94} \\

\cline{2-18}
& \multirow{3}{*}{ LPV~\cite{lpv} } & U14M-Filter  & 98.60 & 97.80 & 98.10 & 89.80 & 93.60 & 97.60 & 95.93 & 86.20 & 78.70 & 75.80 & 80.20 & 82.90 & 81.60 & 82.90 & 81.20  \\
&  & UnionST-S  & 97.77 &	95.98	& 96.85	& 86.47	& 90.85	& 94.44	& 93.73	& 78.44 &	89.85 &	65.00 &	71.63 &	66.65 &	81.31 &	69.25 &	74.59 \\
& & $\Delta$
    & \textcolor{red}{-0.83} & \textcolor{red}{-1.82} & \textcolor{red}{-1.25} & \textcolor{red}{-3.33} & 
      \textcolor{red}{-2.75} & \textcolor{red}{-3.16} & \textcolor{red}{-2.20} & \textcolor{red}{-7.76} & 
      \textcolor{blue}{11.15} & \textcolor{red}{-10.80} & \textcolor{red}{-8.57} & \textcolor{red}{-16.25} & 
      \textcolor{red}{-0.29} & \textcolor{red}{-13.65} & \textcolor{red}{-6.61} \\

\cline{2-18}
& \multirow{3}{*}{ BUSNet~\cite{busnet} } &
U14M-Filter & 98.30 & 98.10 & 97.80 & 90.20 & 95.30 & 96.50 & 96.06 & 83.00 & 82.30 & 70.80 & 77.90 & 78.80  & 71.20 & 82.60 & 78.10  \\
&  & UnionST-S &  97.20	& 95.52 &	96.62 &	85.75	& 91.94 &	94.10 &	93.52 &	73.62 &	87.66 &	63.44 &	69.58	& 62.22 &	79.61 &	68.70 &	72.12 \\
& & $\Delta$  & \textcolor{red}{-1.10}   & \textcolor{red}{-2.58}  & \textcolor{red}{-1.18} & \textcolor{red}{-4.45} & \textcolor{red}{-3.36} & \textcolor{red}{-2.40}  & \textcolor{red}{-2.54} & \textcolor{red}{-9.38} & \textcolor{blue}{5.36} & \textcolor{red}{-7.36} & \textcolor{red}{-8.32} & \textcolor{red}{-16.58} & \textcolor{blue}{8.41} & \textcolor{red}{-13.90} & \textcolor{red}{-5.98}  \\

\cline{2-18}
& \multirow{3}{*}{ CPPD~\cite{cppd} }  & 
U14M-Filter & 99.00 & 97.80 & 98.20 & 90.40 & 94.00 & 99.00 & 96.40 & 86.20 & 78.70 &  76.50 & 82.90 & 83.50 & 81.90 & 83.50 & 81.91 \\
& &  UnionST-S  & 97.17	&95.36	&96.85	&85.15	&91.94	&95.49	&93.66	&83.80	&90.80	&67.11	&75.74	&70.06	&82.04	&69.82	&77.05\\
& & $\Delta$ & \textcolor{red}{-1.83} & \textcolor{red}{-2.44} & \textcolor{red}{-1.35} & \textcolor{red}{-5.25} & \textcolor{red}{-2.06} & \textcolor{red}{-3.51} & \textcolor{red}{-2.74} & \textcolor{red}{-2.40} & \textcolor{blue}{12.10} & \textcolor{red}{-9.39} & \textcolor{red}{-7.16} & \textcolor{red}{-13.44} & \textcolor{blue}{0.14} & \textcolor{red}{-13.68} & \textcolor{red}{-4.86} \\

\midrule
\multirow{15}{*}{ \textbf{AR} } & \multirow{3}{*}{ PARSeq~\cite{parseq} } &
U14M-Filter & 98.90 &98.10 &98.40 &90.10 &94.30 &98.60 &96.40 &87.60 &88.80 &76.50 &83.40 &84.40 &84.30 &84.90 &84.26 \\
& &  UnionST-S  &97.77 &94.74 &97.32 &85.64 &92.09 &93.75 &93.55 &73.54 &88.31 &64.89 &72.66 &68.48 &79.85 &70.24 &74.00 \\
& & $\Delta$ &
\textcolor{red}{-1.13} & \textcolor{red}{-3.36} & \textcolor{red}{-1.08} & \textcolor{red}{-4.46} & \textcolor{red}{-2.21} & \textcolor{red}{-4.85} & \textcolor{red}{-2.85} & \textcolor{red}{-14.06} & \textcolor{blue}{0.49} & \textcolor{red}{-11.61} & \textcolor{red}{-10.74} & \textcolor{red}{-15.92} & \textcolor{blue}{4.45} & \textcolor{red}{-14.66} & \textcolor{red}{-10.26} \\

\cline{2-18}
& \multirow{3}{*}{ MAERec~\cite{jiang2023revisiting} } &
U14M-Filter & 99.20 &97.80 &98.20 &90.40 &94.30 &98.30 &96.36 &89.10 &87.10 &79.00 &84.20 &86.30 &85.90 &84.60 &85.17 \\
&  & UnionST-S  & 98.10 & 96.91 & 98.48 & 88.13 & 95.97 & 95.83 & 95.57 & 78.94 & 90.80 & 68.44 & 77.02 & 71.45 & 83.01 & 73.29 & 77.56 \\
& & $\Delta$ & 
\textcolor{red}{-1.10} & \textcolor{red}{-0.89} & \textcolor{blue}{0.28} & \textcolor{red}{-2.27} & \textcolor{blue}{1.67} & \textcolor{red}{-2.47} & \textcolor{red}{-0.79} & \textcolor{red}{-10.16} & \textcolor{blue}{3.70} & \textcolor{red}{-10.56} & \textcolor{red}{-7.18} & \textcolor{red}{-14.85} & \textcolor{red}{-2.89} & \textcolor{red}{-11.31} & \textcolor{red}{-7.61} \\

\cline{2-18}
& \multirow{3}{*}{ OTE~\cite{ote} } &
U14M-Filter & 98.60 &96.60 &98.00 &90.10 &94.00 &97.20 &95.74 &86.00 &75.80 &74.60 &74.70 &81.00 &65.30 &82.30 &77.09 \\
& &  UnionST-S  &97.77 &94.74 &97.32 &85.64 &92.09 &93.75 &93.55 &73.54 &88.31 &64.89 &72.66 &68.48 &79.85 &70.24 &74.00   \\
& & $\Delta$ &
\textcolor{red}{-0.83} & \textcolor{red}{-1.86} & \textcolor{red}{-0.68} & \textcolor{red}{-4.46} & \textcolor{red}{-1.91} & \textcolor{red}{-3.45} & \textcolor{red}{-2.19} & \textcolor{red}{-12.46} & \textcolor{blue}{12.51} & \textcolor{red}{-9.71} & \textcolor{red}{-2.04} & \textcolor{red}{-12.52} & \textcolor{blue}{14.55} & \textcolor{red}{-12.06} & \textcolor{red}{-3.09} \\

\cline{2-18}
& \multirow{3}{*}{ SMTR~\cite{smtr} } &
U14M-Filter & 99.00 &97.40 &98.30 &90.10 &92.70 &97.90 &95.90 &89.10 &87.70 &76.80 &83.90 &84.60 &89.30 &83.70 &85.00 \\
& & UnionST-S   &97.67 &95.52 &97.78 &86.36 &91.01 &94.10 &93.74 &80.87 &88.31 &65.33 &76.89 &69.24 &84.59 & 71.60 &76.69   \\
& & $\Delta$ &
\textcolor{red}{-1.33} & \textcolor{red}{-1.88} & \textcolor{red}{-0.52} & \textcolor{red}{-3.74} & \textcolor{red}{-1.69} & \textcolor{red}{-3.80} & \textcolor{red}{-2.16} & \textcolor{red}{-8.23} & \textcolor{blue}{0.61} & \textcolor{red}{-11.47} & \textcolor{red}{-7.01} & \textcolor{red}{-15.36} & \textcolor{blue}{4.71} & \textcolor{red}{-12.10} & \textcolor{red}{-8.31} \\

\cline{2-18}
& \multirow{3}{*}{ SVTRv2-AR }   & 
U14M-Filter & 99.03 &97.37 &98.60 &90.56 &95.50 &98.26 &96.56 &91.71 &94.74 &79.44 &86.01 &86.86 &86.29 &85.47 &87.22 \\
& & UnionST-S  & 98.20  & 95.98 & 98.48 & 88.29 & 94.11 & 96.87 & 95.32 & 88.99 & 94.45 & 73.11 & 80.62 & 81.68 & 87.26 & 74.87 & 83.00 \\
& & $\Delta$ &
\textcolor{red}{-0.83} & \textcolor{red}{-1.39} & \textcolor{red}{-0.12} & \textcolor{red}{-2.27} & \textcolor{red}{-1.39} & \textcolor{red}{-1.39} & \textcolor{red}{-1.24} & \textcolor{red}{-2.72} & \textcolor{red}{-0.29} & \textcolor{red}{-6.33} & \textcolor{red}{-5.39} & \textcolor{red}{-5.18} & \textcolor{blue}{0.97} & \textcolor{red}{-10.60} & \textcolor{red}{-4.22} \\

\bottomrule
\end{tabular}}
\caption{Quantitative comparison of different STR models trained on real and synthetic datasets. $\Delta$ denotes the difference between the results obtained on synthetic data and those on real data. \textcolor{red}{Red} indicates a negative value, while \textcolor{blue}{blue} indicates a positive value.}
\label{tab:comp_str_add}
\end{table*}

\section{Model Details}

\textbf{Encoder:} SVTRv2~\cite{du2024svtrv2} begins by partitioning the input image into patches and projects it into a high-dimensional embedding space. It then consists of three stages, each comprising Conv-Blocks and Mixing-Blocks (which use local convolution and global self-attention to capture both local details and global context). Given an input image $\mathbf{I} \in R^{H \times W \times 3}$, the encoder produces a visual feature sequence as follows:
\begin{equation}
    \mathbf{F}_E = \mathrm{Encoder}(\mathbf{I}),
\end{equation}
where $\mathbf{F}_E \in R^{L \times C}$, with $L$ representing the sequence length and $C$ the channel dimension.

\textbf{Decoder:} The AR decoder takes the encoder output $\mathbf{F}_E$ and the embedded target character sequence $\mathbf{T}_{1:t-1}$ (with positional encoding), and models the sequence using two layers of multi-head self-attention and cross-attention:

\begin{equation}
    \mathbf{H}_t = \mathrm{Decoder}( \mathbf{F}_E, \mathrm{Embed}(\mathbf{T}_{1:t-1}) + \mathrm{PE}_{1:t-1})
\end{equation}
where $\mathrm{PE}$ denotes sinusoidal positional encoding. The output at each character step $t$ is computed as:
\begin{equation}
    P(y_t | y_{<t}, \mathbf{F}_E) = \mathrm{Softmax}(\mathbf{H}_t)
\end{equation}

During training, we apply teacher forcing to optimize the cross-entropy loss:
\begin{equation}
    \mathcal{L}_{\mathrm{CE}} = -\sum_{t=1}^{T} \log P(y_t^* | y_{<t}^*, \mathbf{F}_E),
\end{equation}
where $y_t^*$ is the ground-truth character at position $t$. Inference is performed autoregressively.

\section{SEL Details} 
\begin{figure}[ht]
 \centering
 \includegraphics[width=0.9\linewidth]{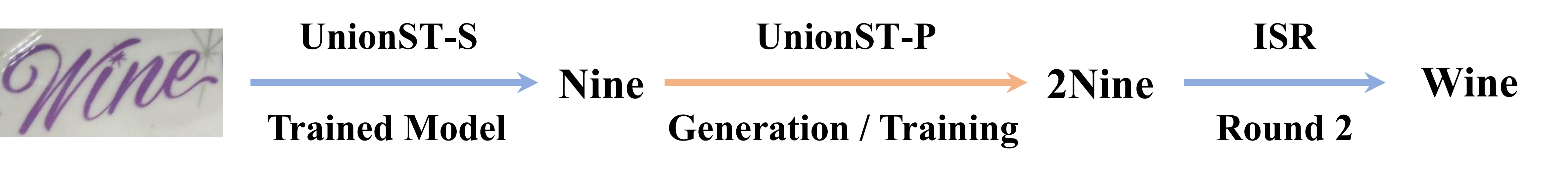}
 \caption{Qualitative examples of intermediate SEL outputs.}
 \label{fig:sel}
\end{figure}

SEL aims to progressively reduce the domain gap between synthetic training data and real-world text by coupling pseudo-label driven corpus construction with iterative self-training. We first train an initial recognizer $\mathcal{M}_a$ on UnionST-S and apply it to a large unlabeled real set $\mathcal{D}_U$ to obtain predictions. The predicted strings $\hat{\mathcal{Y}}_U$ are then used as a target corpus to re-synthesize a new synthetic dataset UnionST-P via the same UnionST rendering engine, yielding UnionST-SP when combined with UnionST-S. Retraining/fine-tuning on UnionST-SP produces a better in-domain model $\mathcal{M}_b$. Starting from $\mathcal{M}_b$, we further perform ISR: at each round $t$, $\mathcal{M}_{t-1}$ is used to pseudo-label $\mathcal{D}_U$, high-confidence samples (above threshold $\tau$, optionally with simple validity checks such as charset/length constraints) are added as $\mathcal{D}_P^{(t)}$ for fine-tuning to obtain $\mathcal{M}_t$, while the remaining low-confidence subset $\mathcal{D}_U^{\text{low}}$ is kept for later manual annotation since it concentrates hard cases (blur, extreme perspective, rare fonts, occlusion, and low contrast). After $T$ rounds, we annotate only $\mathcal{D}_U^{\text{low}}$ to form $\mathcal{D}_L^{\text{hard}}$ and conduct a final fine-tuning step. In the qualitative results in Fig.~\ref{fig:sel}, we provide intermediate examples across rounds (model prediction and the selected status), illustrating how SEL gradually improves recognition on difficult samples, thereby enhancing both training diversity and target-domain robustness.

\section{Training Details} 

Training schedules are adjusted according to dataset size. For synthetic data, we use 40 epochs as the default for 10M UnionST-SP samples. For larger datasets, we sample 10M instances per epoch for 40 epochs (e.g., ST-2D uses a sampling ratio of 0.28). For smaller datasets, the number of epochs is increased proportionally (e.g., 58 epochs for ST). For real data fine-tuning with fewer than 0.1M samples, we train for 100 epochs; for larger real datasets, we use 20 epochs, consistent with training from scratch. For the confidence threshold $\tau$ of ISR, it is set to 0.9. The initial learning rate is set to $6.5 \times 10^{-4}$, while fine-tuning uses $5 \times 10^{-5}$. UnionST synthetic data training employs online resampling augmentation (0.1--1$\times$) and simulates compression loss with a probability of 0.2. Images with a height greater than 1.5 times their width are rotated 90 degrees counterclockwise.

Other settings align with SVTRv2~\cite{du2024svtrv2}: We use the AdamW optimizer~\cite{adamw} with a weight decay of 0.05. The batch size is set to 256. One cycle LR scheduler~\cite{cosine} with 1.5 epochs linear warm-up is used in all epochs, and the maximum text length is set to 25 during training. The character set size is set to 94, including numbers, uppercase and lowercase letters, and common symbols. All models are trained on 8 RTX V100 GPUs. For synthetic-real data mixing, training is first conducted on synthetic data, followed by fine-tuning on real data.

\begin{figure}[t]
\centering
\includegraphics[width=0.48\textwidth]{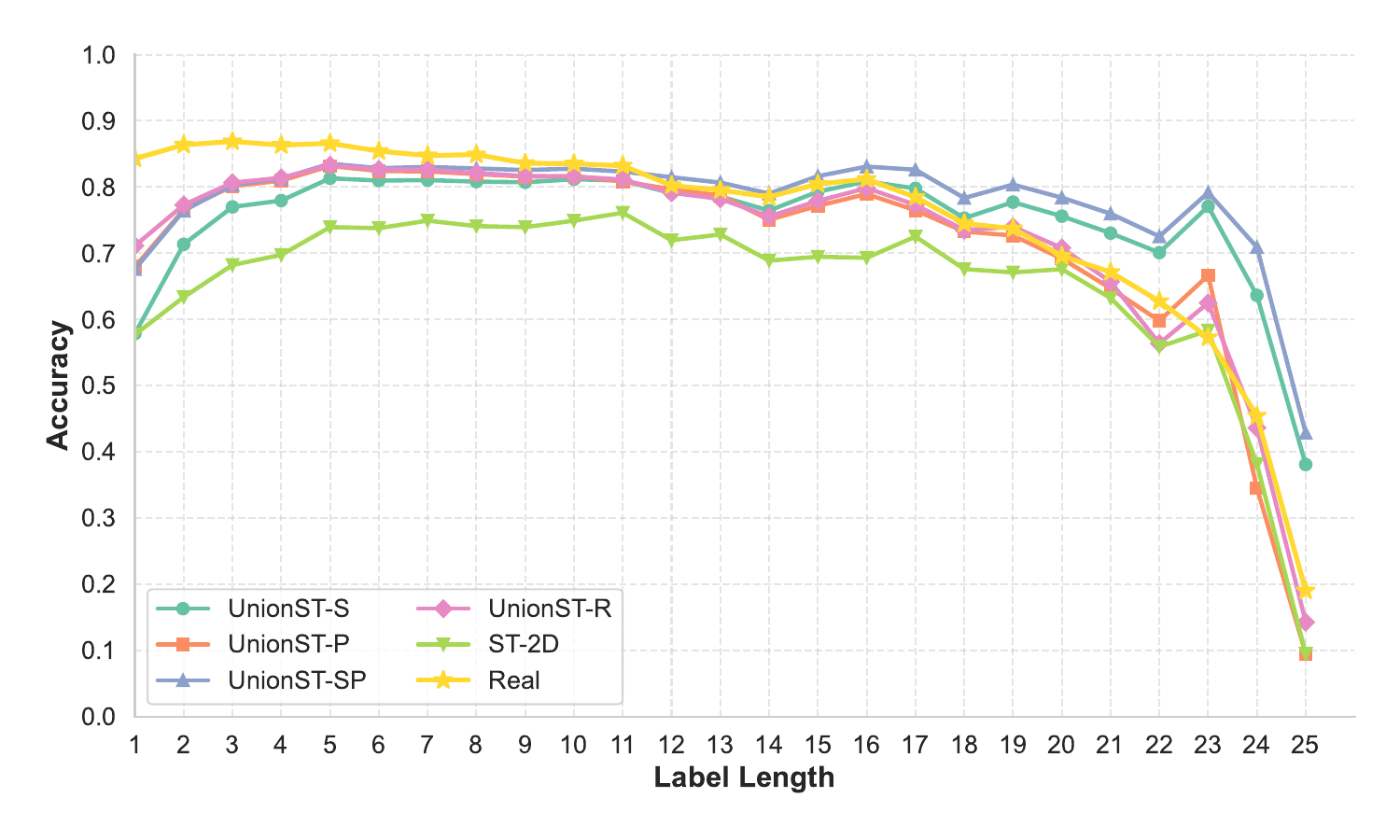} 
\caption{Accuracy as a function of label length on the General subset of the Union14M-Benchmark, evaluated using models trained on different datasets.} 
\label{fig:length_acc}
\end{figure}

\begin{table*}[t]\footnotesize
\centering
\setlength{\tabcolsep}{3pt}{
\begin{tabular}{c|c|ccccccc|cccccccc}
\toprule
\multirow{2}{*}{\textbf{Training Data}} & \multirow{2}{*}{\textbf{Volume}} & \multicolumn{7}{c|}{\textbf{Common Benchmarks}}                                                                & \multicolumn{8}{c}{\textbf{Union14M-Benchmark}}                                             \\
 & &
\textit{IIIT} & \textit{SVT}  & \textit{IC13} & \textit{IC15} & \textit{SVTP} & \textit{CUTE} & \textit{AVG} &
\textit{CUR} & \textit{MLO} & \textit{ART} & \textit{CTL} & \textit{SAL} & \textit{MLW} & \textit{GEN} & \textit{AVG}                  
\\
\midrule

UnionST-S   & 5.00M  & 98.20  & 95.98 & 98.48 & 88.29 & 94.11 & 96.87 & 95.32 & 88.99 & 94.45 & 73.11 & 80.62 & \textbf{81.68} & 87.26 & 74.87 & 83.00 \\

UnionST-S   & 10.0M  & 98.66  & 95.98 & 98.71 & 88.13 & 94.88 & 96.87 & 95.54 & 88.13 & 93.42 & 73.67 & 82.80 & 79.66 & 87.74 & 77.62 & 83.29 \\

UnionST-P   & 5.00M  & 98.70 & 97.22 & 98.60 & 89.23 & 95.97 & 97.57 & 96.21 & 89.45 & 95.03 & 78.22 & 83.83 & 81.49 & 83.50 & 78.58 & 84.30 \\

UnionST-R   & 5.00M  & 98.63 & 96.91 & 98.60 & 89.23 & 95.35 & \textbf{98.61} & 96.22 & 89.28 & 95.47 & 77.33 & 85.62 & 81.43 & 84.95 & 79.39 & 84.78 \\

UnionST-SP  & 10.0M & 98.60 & 97.37 & \textbf{99.07} & 89.29 & 94.88 & 97.22 & 96.07 & 90.64 & \textbf{95.69} & 75.67 & 84.98 & 80.16 & \textbf{87.99} & 78.92 & 84.86 \\

UnionST-SR  & 10.0M & 98.53 & \textbf{97.68} & 98.60 & \textbf{89.67} & 96.28 & 96.87 & 96.27 & 90.48 & 95.03 & 76.67 & 85.37 & 80.99 & 87.50 & 78.98 & 85.00 \\

UnionST-PR  & 10.0M & \textbf{98.87} & 97.06 & 98.72 & 89.40 & \textbf{96.43} & \textbf{98.61} & \textbf{96.51} & \textbf{90.73} & 95.54 & \textbf{79.00} & \textbf{85.88} & 81.49 & 87.38 & \textbf{79.56} & \textbf{85.65} \\

\midrule

UnionST-S~+~R   & 10.0M~+~3.22M    & 99.37	& 98.30 &	\textbf{99.30} &	92.05 &	\textbf{97.83} &	\textbf{99.99} &	97.81 &	94.93 &	97.08 &	85.11 &	88.45 &	90.52 &	91.99 &	87.09 &	90.74  \\

UnionST-SP~+~R   & 10.0M~+~3.22M    & \textbf{99.47} & \textbf{98.92} & \textbf{99.30} & \textbf{92.44} & 96.90 & \textbf{99.99} & 97.84 & \textbf{95.42} & 97.22 & \textbf{86.67} & \textbf{89.60} & \textbf{91.09} & 91.99 & \textbf{87.74} & \textbf{91.39} \\

UnionST-SR~+~R   & 10.0M~+~3.22M    & \textbf{99.47} & 98.76 & \textbf{99.30} & 92.32 & \textbf{97.83} & \textbf{99.99} & \textbf{97.95} & 95.05 & 97.22 & 86.22 & 88.83 & 90.65 & \textbf{92.48} & 87.36 & 91.12 \\

UnionST-PR~+~R   & 10.0M~+~3.22M & \textbf{99.47} & \textbf{98.92} & 99.07 & \textbf{92.44} & 97.36 & \textbf{99.99} & 97.88 & 95.14 & \textbf{97.30} & 86.00 & 89.22 & 90.78 & 91.38 & 87.64 & 91.06 \\

\bottomrule
\end{tabular}}
\caption{Quantitative comparison of UnionST Variants and the experimental setup is consistent with that of the main experiment.}
\label{tab:comp_var}
\end{table*}

\section{More Results}

\subsection{Experiments on Varying Text Lengths}

Fig.~\ref{fig:length_acc} further examines the relationship between label length and recognition accuracy. Across all training datasets, models achieve higher accuracy on shorter labels, with performance declining as label length increases. This trend reflects the increased difficulty of recognizing longer text, due to greater character confusion and error propagation. While models trained on real data perform well on short text, their effectiveness on longer labels is limited by the scarcity of such samples. In contrast, models trained on UnionST-S and UnionST-SP demonstrate greater robustness on longer labels, highlighting the effectiveness of our proposed data for complex and lengthy text recognition.

\subsection{Comparisons with real data on STR models}

Tab.~\ref{tab:comp_str_add} reports the performance gap between synthetic data (UnionST) and real datasets across various STR models. UnionST consistently outperforms real data on Multi-orientated and Multi-words subsets for most models. Notably, the SVTR~\cite{duijcai2022svtr} model achieves a higher average accuracy on UnionST than on real data. Although synthetic datasets still exhibit some differences compared to real data overall, UnionST significantly narrows this gap. Furthermore, UnionST, through its self-evolution semi-supervised framework, reduces reliance on labeled real data while achieving comparable performance.

\subsection{Experiments on UnionST Variants}

The results in Tab.~\ref{tab:comp_var} indicate that, at the same scale, UnionST-SP achieves an average accuracy only 0.12\% lower than the ideal UnionST-SR, which fully leverages both large-scale synthetic and real corpora, on the Union14M-Benchmark. 
Notably, since pseudo-labeling introduces new corpus that closely match the real data distribution, combining these samples with the real dataset enables UnionST-SP to surpass UnionST-SR, yielding a 0.27\% improvement in average accuracy. This further demonstrates the advantages of incorporating pseudo-labeling.

UnionST-PR represents the best performance that UnionST can achieve when the real dataset is known and used for generation. 
Even when the dataset size reaches 10M, the accuracy improvement slows down, revealing the scaling limit, and it still cannot outperform the real dataset. 
It is important to note that UnionST-PR serves as an idealized result, since our SEL framework can't access large-scale real labels. 
Under the setting where the real dataset is already available and used for pre-training, UnionST-SP maintains better performance than UnionST-PR. 
Rather than strictly matching the real distribution, UnionST-SP complements the real dataset, thereby achieving optimal performance. 

\subsection{Backbone-agnostic Gain} Using different backbones all leads to performance gains, we used SVTRv2-AR because it is a strong model, and the pursuit of optimal performance requires both high dataset quality and sufficient model capacity. In addition, we include results showing the gains brought by UnionST-S on other STR models (ABINet~\cite{abinet} and MAERec~\cite{jiang2023revisiting}, see Tab.~\ref{tab:exp_add_bag}), which demonstrate that the improvements are backbone-agnostic.

\begin{table*}[ht]
\footnotesize
\centering
\setlength{\tabcolsep}{3pt}{
\begin{tabular}{c|c|ccccccc|cccccccc}
\toprule
\multirow{2}{*}{\textbf{Model}} & \multirow{2}{*}{\textbf{Training Data}} & \multicolumn{7}{c|}{\textbf{Common Benchmarks}}                                                                & \multicolumn{8}{c}{\textbf{Union14M-Benchmark}}                                             \\
 & &
\textit{IIIT} & \textit{SVT}  & \textit{IC13} & \textit{IC15} & \textit{SVTP} & \textit{CUTE} & \textit{AVG} &
\textit{CUR} & \textit{MLO} & \textit{ART} & \textit{CTL} & \textit{SAL} & \textit{MLW} & \textit{GEN} & \textit{AVG}                  
\\

\midrule
\multirow{3}{*}{ ABINet~\cite{abinet} } &
U14M-Filter & 98.50 &98.10 &97.70 &90.10 &94.10 &96.50 &95.83 &80.40 &69.00 &71.70 &74.70 &77.60 &76.80 &79.80 &75.72 \\
& + UnionST-S &99.10 &98.30	&98.83 &90.89 &94.42 &97.22 &96.46 & 86.31	&92.40	&75.78	&80.62	&82.25	&85.92	&81.04	&83.48
 \\
& $\Delta$ &
\textcolor{blue}{0.60} &  
\textcolor{blue}{0.20} &  
\textcolor{blue}{1.13} & 
\textcolor{blue}{0.79} & 
\textcolor{blue}{0.32} &
\textcolor{blue}{0.72} &
\textcolor{blue}{0.63} &  
\textcolor{blue}{5.91} &  
\textcolor{blue}{23.40} & 
\textcolor{blue}{4.08} & 
\textcolor{blue}{5.92} & 
\textcolor{blue}{4.65} &  
\textcolor{blue}{9.12} &  
\textcolor{blue}{1.24} & 
\textcolor{blue}{7.76} 
\\

\midrule
\multirow{3}{*}{ MAERec~\cite{jiang2023revisiting} } &
U14M-Filter & 99.20 &97.80 &98.20 &90.40 &94.30 &98.30 &96.36 &89.10 &87.10 &79.00 &84.20 &86.30 &85.90 &84.60 &85.17 \\
& + UnionST-S  &99.43	&98.30	&99.30	&92.16	&98.14	&98.61	&97.66	&92.46	&96.71	&83.56	&87.93	&89.07	&91.63	&86.84	&89.74
 \\
 & $\Delta$ &
\textcolor{blue}{0.23} &
\textcolor{blue}{0.50} &
\textcolor{blue}{1.10} &
\textcolor{blue}{1.76} &
\textcolor{blue}{3.84} &
\textcolor{blue}{0.31} &
\textcolor{blue}{1.30} &
\textcolor{blue}{3.36} &
\textcolor{blue}{9.61} &
\textcolor{blue}{4.56} &
\textcolor{blue}{3.73} &
\textcolor{blue}{2.77} &
\textcolor{blue}{5.73} &
\textcolor{blue}{2.24} &
\textcolor{blue}{4.57}
\\

\bottomrule
\end{tabular}
}
\caption{STR gains of UnionST-S by using different backbones.}
\label{tab:exp_add_bag}
\end{table*}

\subsection{Model Predictions Across Training Data}

\begin{figure}[ht]
\centering
\includegraphics[width=0.48\textwidth]{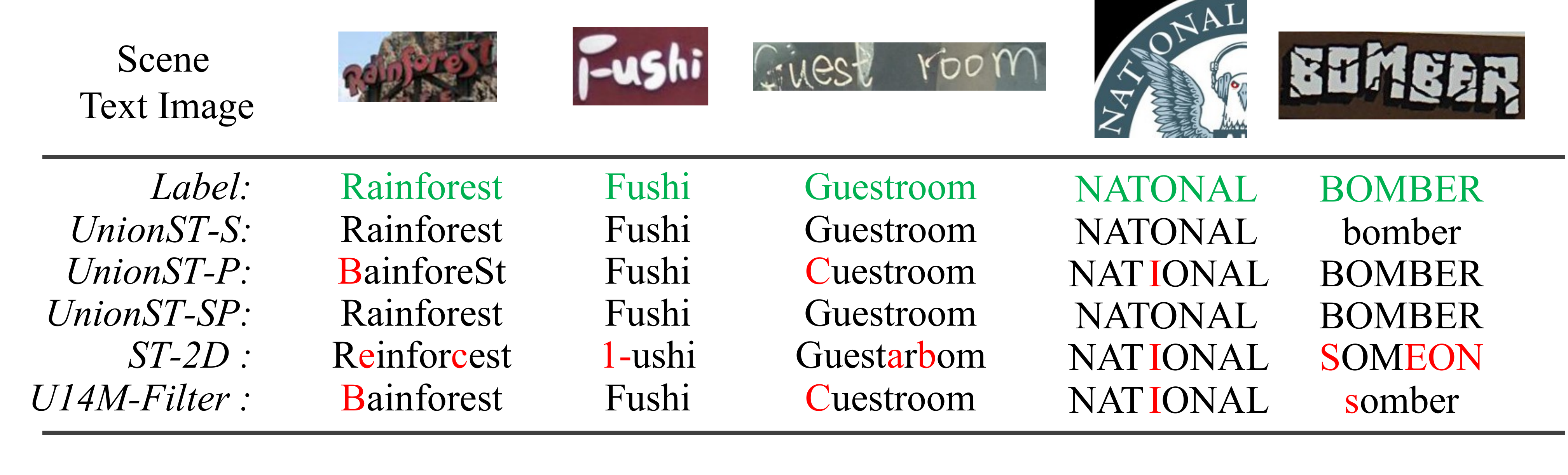} 
\caption{Visualization of model predictions on the Union14M-Benchmark, comparing models trained with different datasets. Characters differing from the ground truth are highlighted in red.} 
\label{fig:com_vis}
\end{figure}

As shown in Fig.~\ref{fig:com_vis}, predictions from models trained on UnionST are generally consistent with the ground truth. Notably, when the model trained on real data produces errors, UnionST-P often replicates these mistakes. However, UnionST-S frequently corrects such errors, underscoring the benefit of integrating both approaches in UnionST-SP. This combined strategy achieves performance comparable to models trained on real data, while mitigating certain biases inherent in real datasets. For instance, in the ``NATONAL” example, the letter ``I” is occluded. Both the real dataset and UnionST-P tend to hallucinate the missing character, whereas UnionST-S and UnionST-SP adhere strictly to the visual evidence and do not insert the absent letter.

\subsection{Hard Cases}

\begin{figure}[ht]
\centering
\includegraphics[width=0.48\textwidth]{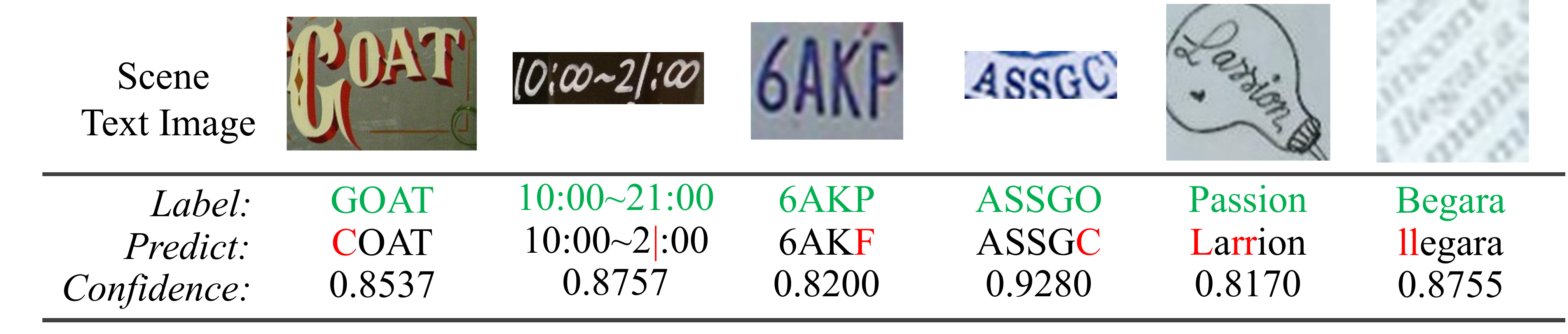} 
\caption{Examples of incorrect predictions by the best-performing model (average accuracy: 91.39\% on Union14M-Benchmark).} 
\label{fig:hard}
\end{figure}

Fig.~\ref{fig:hard} presents representative failure cases, which can be categorized as follows: (1) \textbf{Font-induced character similarity}: Artistic fonts can obscure distinctions between characters, as in the first example where ``G” and ``C” are easily confused, or in the fifth example where ``L” resembles ``P”.  (2) \textbf{Visual-semantic trade-off}: Some characters, such as ``$|$” and ``1”, are visually ambiguous and require contextual understanding. Modeling such semantics about time remains challenging due to the scarcity of relevant training data. (3) \textbf{Ambiguity from occlusion or cropping}: In the third example, ``P” is misrecognized as ``F” due to partial occlusion, and the lack of context makes both interpretations plausible. (4) \textbf{Label noise}: The fourth example demonstrates a labeling error, where the ground truth should be ``ASSGC” instead of ``ASSGO”, causing the correct prediction to be marked as incorrect. (5) \textbf{Extremely challenging cases}: In the sixth example, severe image blur renders the text nearly illegible, making accurate recognition difficult even for human annotators.

\section{Discussion}
\subsection{Emulation of Real-world} Even with synthetic backgrounds, we can create highly valuable training data that effectively emulates real-world variability and applicability, as demonstrated by the improved results in Tab.~\ref{tab:comp}. 
For example, occlusions is modeled with the ``Incomplete'' operation, and multi-line text shares the same pipeline with vertical text (a special case of multi-line layout). 
We acknowledge that there are still differences compared to fully realistic scene images. However, purely real data is hard to balance and scale, and combining real and synthetic data becomes a clear trend to the OCR community. Then, the SEL further narrows the synthetic gap in real scenes while requiring only limited human effort. Moreover, UnionST can serve as a practical foundation for scalable, labeled full-scene synthesis, e.g., it first produces accurately labeled ``sketches'' with diverse layouts, then generative models refine these into photorealistic full scenes. 

\subsection{Multilingual Support} 
\begin{figure}[ht] \centering
  \includegraphics[width=0.48\textwidth]{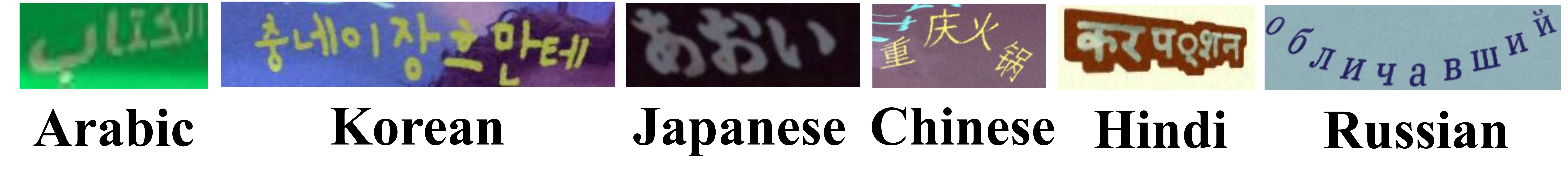}
    \vspace{-2.5em}
    \caption{Visualization of synthesized multilingual examples.} \label{fig:mlt}
\end{figure}

UnionST inherently supports multiple languages, and it can perform data synthesis as long as the corresponding language corpus is provided. As illustrated in Fig.~\ref{fig:mlt}, UnionST is capable of generating multilingual examples with significant diversity and realism. In our next steps, we will develop a multilingual version of UnionST. We also plan to construct a similar multilingual evaluation benchmark following the ``U14M-Bench'' creation protocol, which covers diverse and challenging scenarios, to assess the utility of UnionST.

\subsection{Consideration for Using SEL} Our goal is to verify the capability of the UnionST data engine. Therefore, we adopt SEL, a simple yet effective self-/semi‑supervised learning baseline. Even in this straightforward setting, UnionST already works effectively and substantially reduces the annotation cost. UnionST can be seamlessly integrated with other self-/semi‑supervised methods and we believe that incorporating more advanced methods would further improve the performance. Exploring such combinations is an important direction we plan to pursue.

\end{document}